\documentclass{article}
\usepackage{tabularx}

\usepackage[final]{corl_2026} 

\title{AURORA: Active Uncertainty-Driven Re-Orientation for In-Hand Reconstruction}

\author{
Feiyu Zhao$^*$\\
  School of Information Science\\ and Technology\\
  ShanghaiTech University, China\\
  \texttt{zhaofy12024@shanghaitech.edu.cn}
  \And
Yuetong Li$^*$\\
  School of Information Science\\ and Technology\\
  ShanghaiTech University, China\\
  \texttt{liyt2023@shanghaitech.edu.cn}
  \And
Chenxi Xiao$^\dagger$\\
  School of Information Science\\ and Technology\\
  ShanghaiTech University, China\\
  \texttt{xiaochx@shanghaitech.edu.cn}
}

\usepackage{amsmath,amssymb, amsfonts}
\usepackage{algorithm}
\usepackage{array}
\usepackage{textcomp}
\usepackage{stfloats}
\usepackage{url}
\usepackage{verbatim}
\usepackage{graphicx}

\usepackage{svg}
\usepackage{xcolor}
\usepackage{bm}
\usepackage{booktabs}
\usepackage{multirow}
\usepackage{algpseudocode}
\usepackage[labelfont=bf]{caption}
\usepackage{subcaption} 
\usepackage{makecell}
\usepackage{stfloats}
\usepackage{float}
\hypersetup{
  colorlinks = true,
  urlcolor   = blue,
  linkcolor  = blue,
  citecolor  = blue
}

\usepackage{tikz}
\usetikzlibrary{calc,positioning,arrows.meta}
\usepackage{placeins}
\usepackage{siunitx}
\begin{document}
\maketitle

\begingroup
\renewcommand{\thefootnote}{}
\footnotetext{$^*$Equal contribution. $^\dagger$Corresponding author.}
\endgroup

\begin{center}
    \centering
    \captionsetup{type=figure}
    \includegraphics[width=0.9\textwidth]{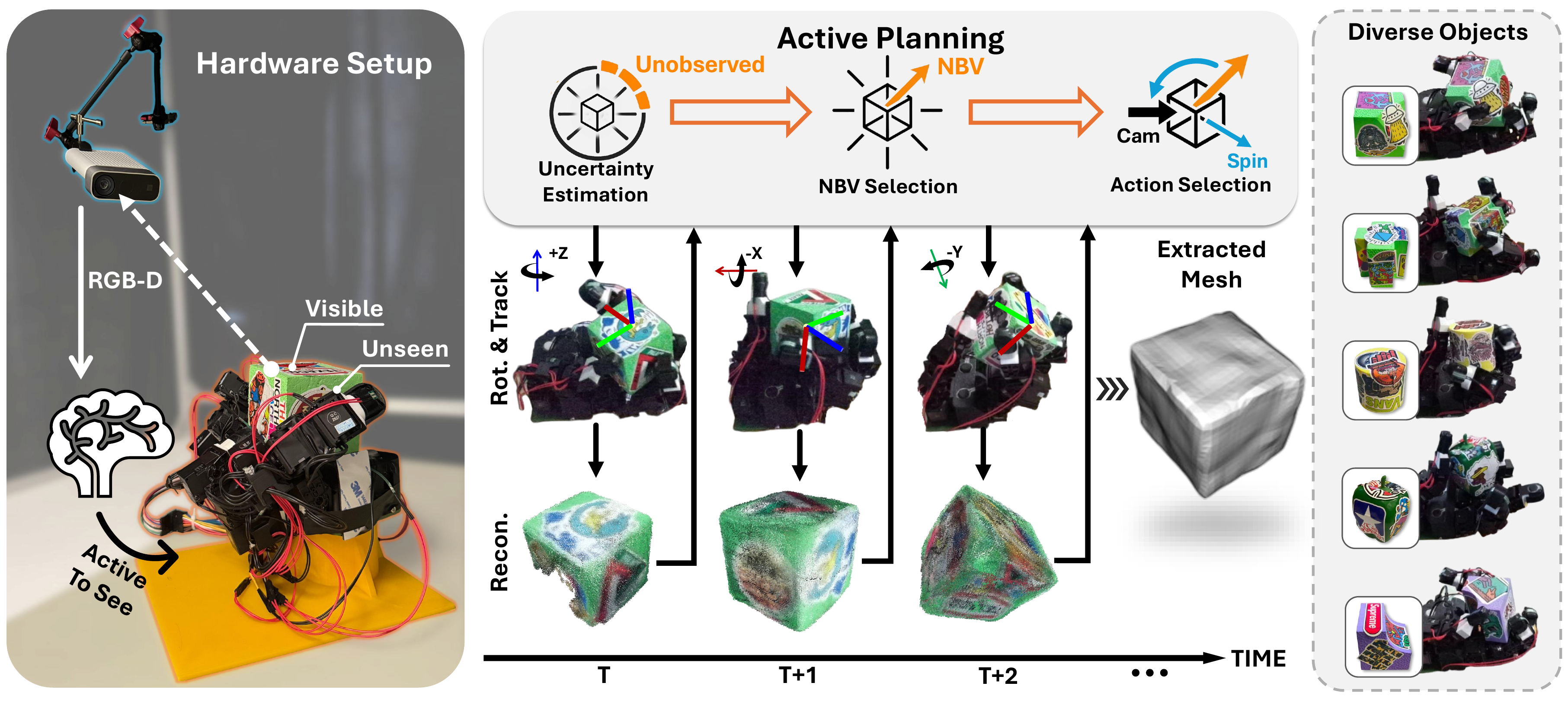}
    \captionof{figure}{AURORA, a framework for active in-hand object reconstruction using a fixed RGB-D camera. The hand reorients an object using an in-hand manipulation policy and an uncertainty-driven next-best-view planner, exposing unobserved regions and accelerating 3D reconstruction.}
\end{center}%


\begin{abstract}
Observing objects grasped by a robot hand is challenging due to severe visual occlusions. Although in-hand manipulation can expose hidden surfaces, existing approaches often rely on predefined or open-loop reorientation strategies that do not explicitly target under-observed regions. We propose AURORA, an active 3D reconstruction framework that closes the loop between online object-centric reconstruction and in-hand reorientation. At its core, Ray-GPIS estimates direction-wise reconstruction uncertainty along candidate viewing rays and selects next-best-view targets using an uncertainty--novelty objective, which are realized through an axis-conditioned in-hand rotation policy. The resulting RGB-D observations are fused incrementally using CAD-free 6D pose tracking and lightweight geometric reconstruction. Experiments demonstrate that AURORA improves reconstruction quality and information-acquisition efficiency over non-active rotation strategies, while Ray-GPIS also outperforms active view-planning baselines in reconstruction performance, action-ranking quality, and planning efficiency. Targeted ablations further validate its robustness to hand occlusion and pose errors. The project webpage is available at \url{https://aurorahand.github.io/}.

\end{abstract}

\keywords{Probabilistic learning and uncertainty in robotics,  Multimodal perception, sensor fusion, and robot vision, Active In-Hand Object Reconstruction
} 


\section{Introduction}
	
Human perception is inherently active: when inspecting an object, humans deliberately reorient it to reveal surfaces occluded by the hand~\citep{bajcsy1988active}. Robotic systems have a similar need during manipulation, where hand-object occlusions and limited camera viewpoints lead to partial observability. Unlike humans, who can simultaneously coordinate vision and manipulation to acquire novel views, most robotic systems rarely leverage their physical dexterity to resolve such occlusions, restricting the information available for understanding the object.

A primary reason for this limitation is the lack of a planning framework that translates perceptual needs into manipulation actions. While prior research has explored using robot manipulation data for object understanding, these approaches typically reconstruct objects from passively collected manipulation sequences, such as pick-and-place interactions~\citep{krainin2011manipulator} or in-hand rotation videos~\citep{suresh2024neuralfeels}. In these methods, the robot follows predefined action sequences or open-loop policies, serving merely as a passive data collector without determining where to observe next. Because they do not explicitly reason about viewpoint informativeness or convert perceptual uncertainty into action decisions, such pipelines are time-inefficient and may leave occluded regions persistently under-observed.

To efficiently and accurately recover the complete 3D geometry of unknown objects, we propose an active in-hand reconstruction framework that tightly couples visual perception with manipulation in a closed-loop manner. Rather than executing predefined motion sequences, the system follows an informative policy that actively selects actions to improve perceptual coverage. At each time step, the robot explicitly represents spatial geometric uncertainty over the object surface, which is then used to plan the next in-hand reorientation action, accelerating the exposure of under-observed regions. On the basis of the collected views, we implement a geometric fusion framework to integrate multi-view observations. Together, these components enable fully autonomous object reconstruction through coordinated hand–eye interactions.

In summary, the contributions of this paper include:
\begin{itemize}
    \item AURORA, an informative framework for capturing complete object geometry in-hand, enabling autonomous exploration with improved efficiency and completeness.
    \item {Ray-GPIS, a ray-conditioned planner that converts GPIS point-wise uncertainty into direction-wise view scores for active in-hand reorientation.}
    \item {
    An online object-centric fusion pipeline, and extensive real-world evaluations showing improved completeness and exploration efficiency under hand-object occlusion.}
\end{itemize}

\section{Related Works}\label{sec:related}

\subsection{In-Hand Manipulation for Object Reconstruction}

In-hand manipulation enables a robot to reorient an object within its grasp, providing a natural mechanism for exposing previously unseen surfaces. Existing approaches have studied both model-based manipulation, which plans motions using contact mechanics~\citep{chavan2018hand, liang2024robust}, and learning-based manipulation, which improves robustness to novel objects and complex dynamics~\citep{qi2022rapid, khandate2023sampling, openai2018learningdex, chen2022visualdexterity, qin2022dexpoint}. Recent works further incorporate tactile sensing to improve contact perception and dexterous control~\citep{yin2023rotating, yuan2023robotsynesthesia, qin2026nlipscalib}.

However, most in-hand manipulation methods optimize for reorientation stability, speed, or task success, rather than reconstruction quality. A recent work NeuralFeels~\citep{suresh2024neuralfeels} is closely related, as it reconstructs in-hand object geometry from visuotactile observations. 
Despite this similarity, NeuralFeels reconstructs object pose and shape from a given sequence of visual and high-resolution tactile interactions, but does not include an active planning mechanism. Although a direct benchmark is not feasible due to differences in hardware settings, we discuss the relationship between NeuralFeels and our work along with limited comparisons in Appendix~\ref{app:neuralfeels_comparison}.

\subsection{Object Reconstruction and Tracking under Occlusion}

Object reconstruction from image sequences has been widely studied, with approaches ranging from Structure-from-Motion~\citep{schoenberger2016sfm} and RGB-D fusion to neural implicit representations~\citep{lei2020pix2surf, munkberg2022extracting, sun2021neuralrecon, azinovic2022neural}. 
These methods usually assume static scene geometry and establish multi-view consistency primarily through camera motion. 
In contrast, in robotic in-hand object reconstruction, the object moves unpredictably under contact and is occluded by the hand, making such multi-view consistency more difficult to achieve. 
Recent model-free trackers and foundation-model-based pose estimators provide useful tools for addressing this issue~\citep{wen2021bundletrack, wen2024foundationpose, lee2025any6d, liu2025one2any}, and have been used to address similar occlusion challenges arising in human--object interaction reconstruction~\citep{jiang2025dynhor, huang2022intercap, jiang2022neuralhofusion, cao2021reconstructing}. 
However, these works primarily reconstruct objects from observed sequences and do not address active action or viewpoint selection for revealing previously unobserved object regions.

\subsection{Active Exploration and Uncertainty-Driven View Planning}

Active perception seeks to plan sensing actions that maximize task-relevant information acquisition. Most existing systems explore static, unknown environments using mobile sensors~\citep{cao2020hierarchical, feng2024fc, zhou2021fuel}. These methods use sensing modalities such as cameras or tactile sensors~\citep{zhao2025autonomous, xiao2022active} and typically optimize objectives that reduce geometric uncertainty or maximize expected information gain~\citep{stachniss2005information}. Common formulations include frontier-based exploration of unknown space~\citep{yamauchi1997frontier}, as well as information-theoretic criteria based on Fisher information~\citep{jiang2024fisherrf}, mutual information~\citep{xie2025gauss}, and related measures. Beyond mobile-camera exploration, recent work has also studied active object reconstruction with fixed sensors and manipulable objects. For instance, a robot can grasp and re-pose an object in front of a static RGB-D camera to expose previously occluded surfaces~\citep{krainin2011autonomous}. Other interactive perception approaches use purposeful manipulation to incrementally reveal unseen geometry and improve reconstruction completeness~\citep{bohg2017interactive}. However, these methods largely assume controlled object reposing or external manipulation, rather than continuous in-hand manipulation. As a result, they do not address the coupled challenges of persistent hand-induced occlusion, object motion during interaction, and action selection for revealing uncertain object regions in the robot hand.


\section{Methodology}
\begin{figure}[t]
    \centering
    \includegraphics[width=0.90\textwidth]{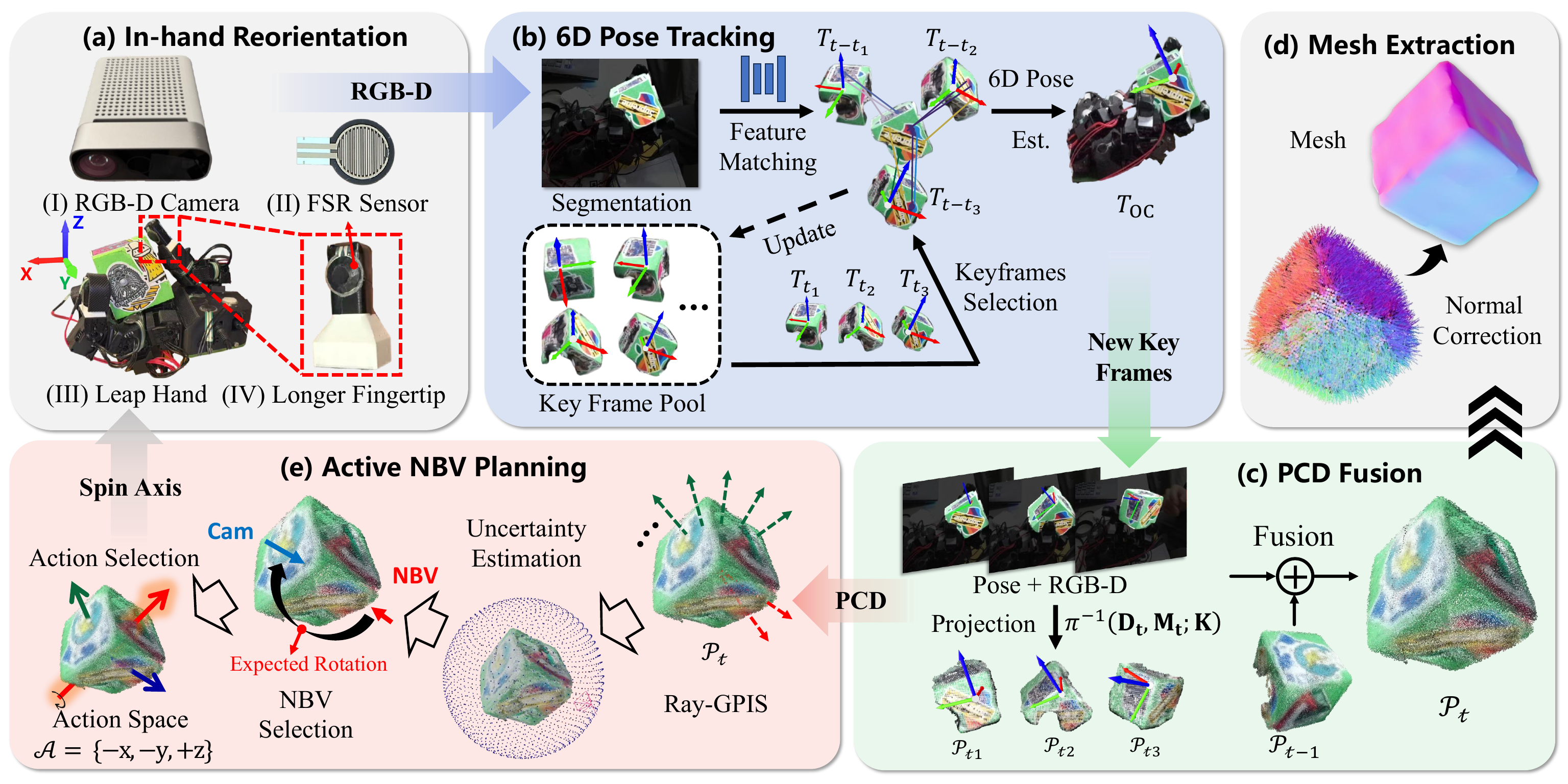}
    \caption{Overview of our technical pipeline. The system integrates four modules: (a) in-hand object reorientation with the Leap Hand; (b) 6D pose tracking via BundleTrack \citep{wen2021bundletrack}; (c--d) reconstructions; and (e) uncertainty-driven next-best-view planning.}
    \label{fig:pipeline}
    \vspace{-0.3cm}
\end{figure}

The overall pipeline of AURORA is illustrated in Fig.~\ref{fig:pipeline}. We tackle in-hand 3D reconstruction as a closed-loop perception--action problem. From an initial observation, we iterate: (i) select a rotation axis and execute an in-hand rotation; (ii) track the object pose and select keyframes; (iii) fuse keyframes into an incremental point-cloud reconstruction; and (iv) update an uncertainty model and compute the next-best view (NBV) for the next action. When the loop terminates, we extract a watertight mesh as the final reconstruction.

\subsection{In-Hand Object Reorientation}
\label{subsec:method_inhand_rotation}

We use a tactile-based in-hand rotation policy adapted from Rotating without Seeing~\citep{yin2023rotating} as the low-level action executor. Given a commanded rotation axis, the policy reorients the grasped object accordingly. We deploy the policy on a Leap Hand~\citep{shaw2023leaphand}; hardware modifications, tactile sensing, and policy training details are provided in Appendix~\ref{app:inhand_policy_details}.

We define our action space in a hand-centered world frame \(W\), whose origin is at the palm center. The \(z\)-axis aligns with the outward palm normal, and the \(x\)- and \(y\)-axes lie in the palm plane. The feasible action set is:
\begin{equation}
\mathcal{A} \triangleq \{-x,\ -y,\ +z\},
\label{eq:action_set}
\end{equation}
where each element denotes a rotation-axis primitive in \(W\). At each planning step, the active planner selects \(\mathbf{a}^{\star}\in\mathcal{A}\), which is then executed by the low-level policy.

\subsection{Ray-GPIS Active View Planning}
\label{subsec:method_active_gpis}

\begin{figure}[t]
    \centering
    \includegraphics[width=0.90\linewidth]{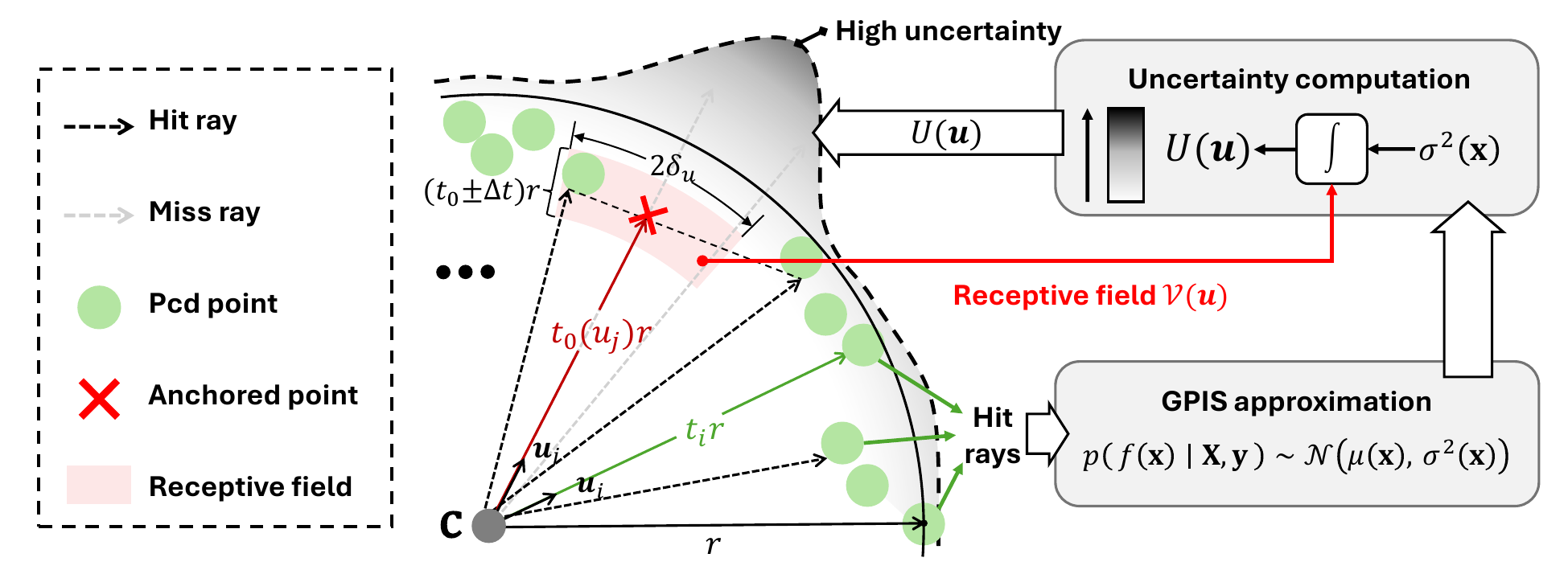}
    \caption{
    Ray-GPIS estimates direction-wise reconstruction uncertainty. Rays are cast from the object center \(\mathbf{c}\). Each direction \(\mathbf{u}\) is assigned an anchor point \(\mathbf{x}_0(\mathbf{u})\) near the predicted surface intersection, and GPIS variance is aggregated in a local receptive field to obtain \(U(\mathbf{u})\).
    }
    \label{fig:raygpis_schematic}
\end{figure}

Actions are selected based on reconstruction uncertainty over object geometry.
While existing GPIS-based methods~\citep{williams2006gaussian, zhao2025autonomous} estimate point-wise uncertainty in 3D space, they do not directly provide view-dependent scores for in-hand exploration.
To address this, Ray-GPIS converts point-wise spatial GPIS uncertainty into direction-wise planning scores by casting object-centered rays, anchoring them near predicted surface locations, and aggregating posterior variance within local receptive fields, as shown in Fig.~\ref{fig:raygpis_schematic}.

\subsubsection{Ray-GPIS Uncertainty Estimation}

Specifically, we use the center \(\mathbf{c}\) of the oriented bounding box of \(\mathcal{P}_t\) as the ray origin and uniformly sample \(N\) unit directions \(\{\mathbf{u}_i\}_{i=1}^{N}\) on the sphere. 
Each ray is parameterized by a normalized radial coordinate \(t_i\):
\begin{equation}
\mathbf{x}(t_i;\mathbf{u}_i) = \mathbf{c} + t_i r \mathbf{u}_i, \qquad t_i\in[0,t_{\max}],
\label{eq:ray_parameterization}
\end{equation}
where \(r\) is the maximum distance from \(\mathbf{c}\) to any point in \(\mathcal{P}_t\), and \(t_i\) denotes the radial distance normalized by \(r\). Thus, \(t_i=1\) approximately corresponds to the current point-cloud extent along the object scale, while \(t_{\max}>1\) allows the ray to probe slightly beyond the observed geometry.

For each direction, Ray-GPIS assigns an anchor point
\(\mathbf{x}_0(\mathbf{u}_i)=\mathbf{x}(t_0(\mathbf{u}_i);\mathbf{u}_i)\),
which approximates the surface location along the ray.
A ray is classified as a \emph{hit ray} if it intersects the observed surface neighborhood represented by \(\mathcal{P}_t\); its anchor is placed at the first observed surface intersection.
A \emph{miss ray} has no reliable observed intersection.
For miss rays, the anchor depth is inferred by interpolating nearby hit-ray depths on the viewing sphere.
This enables Ray-GPIS to query uncertainty both around observed surfaces and near plausible surface locations in unobserved directions.
Details of the definitions and implementation are provided in Appendix~\ref{app:ray_gpis_definition}.

We then fit a GPIS~\citep{williams2006gaussian} model to the observed anchors and use its posterior variance \(\sigma^2(\mathbf{x})\) as the spatial reconstruction uncertainty.
The GPIS is trained only on hit-ray anchors, while miss-ray anchors are used only as query locations for uncertainty evaluation.
To obtain a direction-wise score, we evaluate the variance within a local receptive field around each anchor:
\begin{equation}
\mathcal{V}(\mathbf{u}_i)
=
\left\{
\mathbf{x}(t_i;\mathbf{u})
\,\middle|\,
t_i\in[t_0(\mathbf{u}_i)-\Delta t,\ t_0(\mathbf{u}_i)+\Delta t],
\ \angle(\mathbf{u},\mathbf{u}_i)\le \delta_u
\right\}.
\label{eq:receptive_field}
\end{equation}
Here, \(\Delta t\) defines a narrow radial band around the anchor depth, and \(\delta_u\) defines the angular neighborhood on the viewing sphere.

The direction-wise uncertainty is computed by integrating the GPIS variance over the receptive field:
\begin{equation}
U(\mathbf{u}_i)
=
\int_{\mathcal{V}(\mathbf{u}_i)} \sigma^2(\mathbf{x})\, d\mathbf{x},
\label{eq:rf_uncertainty}
\end{equation}
Details of GPIS training and posterior inference are provided in Appendix~\ref{app:train_gpis}.

\subsubsection{Next-Best-View Selection}
\label{subsec:method_active_nbv}

Given the direction-wise uncertainty \(U(\mathbf{u}_i)\) estimated by Ray-GPIS, the next view should target regions that are both uncertain and insufficiently observed. 
However, directly maximizing \(U(\mathbf{u}_i)\) may select directions close to the current fused geometry, leading to redundant observations. 
We therefore introduce a novelty weight based on the distance between each candidate receptive field and the current point cloud \(\mathcal{P}_t\). 
For a candidate direction \(\mathbf{u}_i\), let \(\bar{d}_{\mathrm{nn}}(\mathbf{u}_i)\) be the average nearest-neighbor distance from samples in \(\mathcal{V}(\mathbf{u}_i)\) to \(\mathcal{P}_t\). 
We normalize it by the median 1-NN spacing \(\bar{d}_{1\mathrm{nn}}\) of \(\mathcal{P}_t\), which represents the typical point-cloud resolution, and define
\begin{equation}
\eta(\mathbf{u}_i)
=
1-\exp\!\left(
-\left(
\frac{\bar{d}_{\mathrm{nn}}(\mathbf{u}_i)}
{\beta \bar{d}_{1\mathrm{nn}}}
\right)^2
\right),
\label{eq:novelty_weight}
\end{equation}
where \(\beta\) controls the scale at which a direction is considered novel. 
The final NBV direction is selected by maximizing the product of reconstruction uncertainty and geometric novelty:
\begin{equation}
\mathbf{u}^{\star}_{\mathrm{NBV}}
=
\arg\max_{\mathbf{u}_i}
U(\mathbf{u}_i)\eta(\mathbf{u}_i).
\label{eq:nbv_selection}
\end{equation}
This encourages the planner to prioritize regions that are not only uncertain, but also far from already fused observations.

\subsubsection{Action Selection}
\label{subsec:method_active_axis}

The selected NBV direction specifies which object-side region should be exposed to the fixed camera, but it must still be converted into an executable in-hand rotation primitive. 
Since the low-level controller operates in the hand-centered world frame \(W\), we first transform the object-frame NBV direction into \(W\):
\begin{equation}
\mathbf{d}_{W}
=
\mathbf{R}_{WO}(t)\mathbf{u}^{\star}_{\mathrm{NBV}},
\label{eq:nbv_transform_world}
\end{equation}
where \(\mathbf{R}_{WO}(t)\) is the rotation component of \(\mathbf{T}_{WO}(t)=\mathbf{T}_{WC}\mathbf{T}_{CO}(t)\). 
We define \(\mathbf{v}_W\) as the unit vector from the current object center to the camera, expressed in \(W\).
We compute the minimal object rotation that aligns the desired direction \(\mathbf{d}_{W}\) with \(\mathbf{v}_{W}\), and represent it by the Lie-algebra vector \(\boldsymbol{\omega}\in\mathbb{R}^{3}\). 
Since the controller only supports a discrete action set \(\mathcal{A}\), we choose the primitive whose axis is most aligned with this desired rotation:
\begin{equation}
\mathbf{a}^{\star}
=
\arg\max_{\mathbf{a}\in\mathcal{A}}
\mathbf{a}^{\top}\boldsymbol{\omega}.
\label{eq:axis_selection}
\end{equation}
Details of the frame transformation and rotation-vector computation are provided in Appendix~\ref{app:action_selection_details}.

\subsection{Object-Centric Tracking and Reconstruction}
\label{subsec:method_tracking_keyframe}
\label{subsec:method_online_recon}

Object-centric fusion and Ray-GPIS planning require object poses in RGB-D frames. 
We use BundleTrack~\citep{wen2021bundletrack}, a CAD-free RGB-D tracker, to estimate \(T_{CO}(t)\in SE(3)\) from segmented observations~\citep{carion2025sam3segmentconcepts} and fuse them in a consistent object frame. 
To reduce the effect of hand--object occlusions, we apply a lightweight visibility-aware keyframe filter that retains only frames with sufficient object visibility, stable pose estimates, and adequate motion from the latest keyframe. 
Detailed tracking and keyframe-selection rules are provided in Appendix~\ref{app:tracking_details}.

Given the selected keyframes \(\mathcal{K}\), we maintain a compact object-centric point cloud as the geometric state for closed-loop planning. 
For each keyframe \(t\in\mathcal{K}\), masked depth pixels are back-projected into the camera frame, transformed into the object frame, and fused incrementally:
\begin{equation}
\mathcal{P}_t \triangleq
\mathcal{C}\!\left(
\mathcal{P}_{t-1}\ \cup\
T_{CO}(t)^{-1}
\pi^{-1}\!\big(\mathbf{D}_t,\mathbf{M}_t;\mathbf{K}_{\mathrm{cam}}\big)
\right),
\label{eq:online_fusion_compact}
\end{equation}
where \(\mathbf{D}_t\), \(\mathbf{M}_t\), and \(\mathbf{K}_{\mathrm{cam}}\) denote the depth image, object mask, and camera intrinsics, respectively; \(\pi^{-1}(\cdot)\) back-projects masked depth pixels into 3D camera-frame points; and \(\mathcal{C}(\cdot)\) applies voxel downsampling and outlier removal. 
The resulting point cloud \(\mathcal{P}_t\) is used by Ray-GPIS for uncertainty estimation and NBV planning.

After exploration, we convert the final fused point cloud \(\mathcal{P}_T\) into a watertight mesh for quantitative evaluation. 
We estimate globally consistent normals using FaCE~\citep{scrivener2025faraday} and reconstruct the mesh with NKSR~\citep{huang2023nksr}; details are provided in Appendix~\ref{app:mesh_extraction_details}.

\section{Experiments}
\label{sec:experiment}

We evaluate the proposed active in-hand reconstruction framework on six graspable real-world objects with varied geometries. Section~\ref{subsec:exp_setup} describes the hardware setup and experimental protocol. Section~\ref{subsec:exp_recon_quality} evaluates reconstruction accuracy. Section~\ref{subsec:exp_baseline_comparison} compares AURORA with both non-active manipulation strategies and active view-planning baselines, evaluating reconstruction efficiency and action-selection quality. Finally, Sec.~\ref{subsec:exp_ablation} analyzes the contributions and robustness of the key Ray-GPIS components through targeted ablations.

\subsection{System Setup and Task Protocol}
\label{subsec:exp_setup}

Our system uses a fixed Azure Kinect DK RGB-D camera for visual sensing and a Leap Hand~\citep{shaw2023leaphand} for in-hand manipulation. 
During execution, the hand rotates the object for \(6\,\mathrm{s}\); the planner then updates the reconstruction and replans the next rotation direction.
This closed-loop process is repeated under a fixed \(30\,\mathrm{s}\) interaction budget. Then, we evaluate both the online point-cloud reconstruction used for active planning and the final mesh extracted after exploration. 
The \(6\,\mathrm{s}\) interval is determined by the low-level hand controller,
which requires this time to complete the commanded rotation and stabilize the
grasp, rather than by planner computation. Detailed module runtimes are
reported in Appendix~\ref{app:runtime}.
For quantitative evaluation, we obtain ground-truth meshes using an EinScan Pro 2X high-resolution 3D scanner.
More detailed hardware specifications and hyperparameters for the planner and reconstruction are provided in Appendix~\ref{app:hardware_specification} and \ref{app:experiment_parameters}, respectively.

\subsection{Reconstruction Accuracy}
\label{subsec:exp_recon_quality}

\begin{figure}[t]
    \centering
    \includegraphics[width=0.9\textwidth]{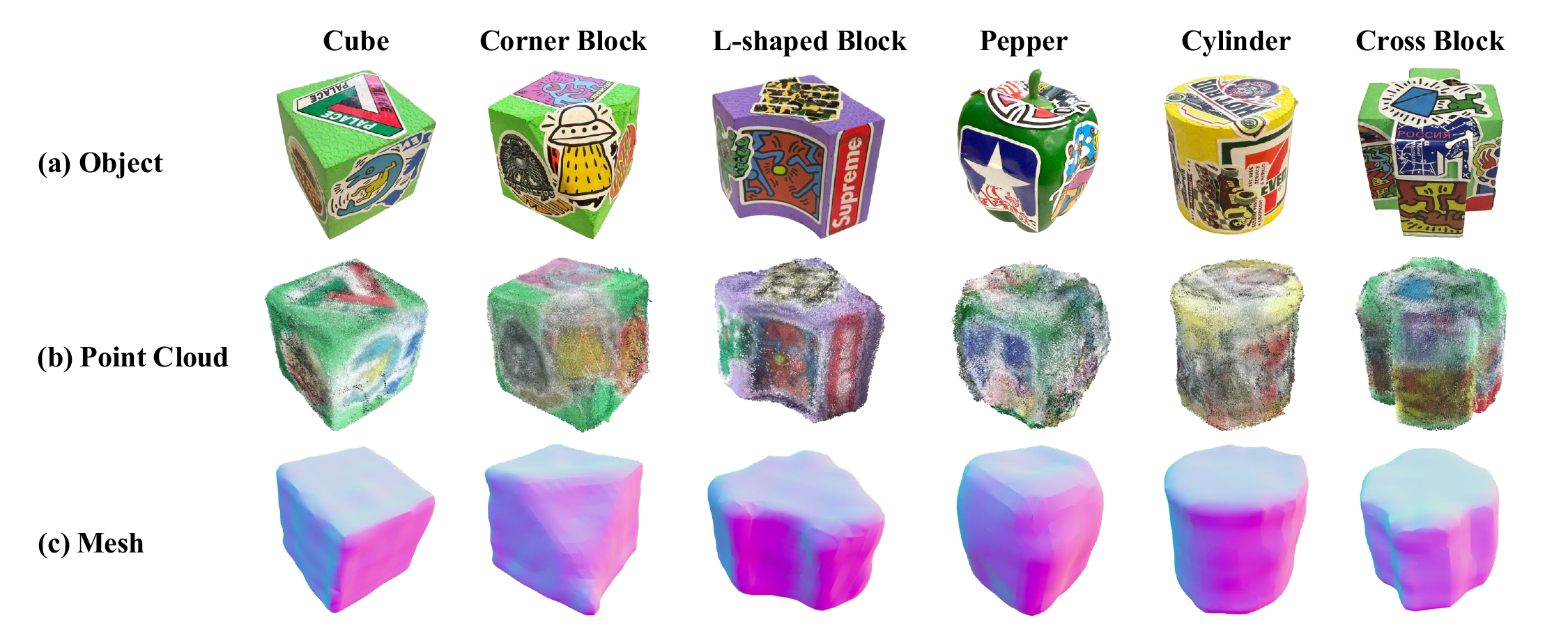}
    \caption{
    Qualitative results on real-world objects.
    (a) Test objects.
    (b) Online point clouds.
    (c) Offline extracted meshes. 
    }
    \label{fig:qual_all_objects}
\end{figure}

The first evaluation focuses on the reconstruction quality of the proposed AURORA framework. We evaluate reconstruction performance on six graspable real-world objects. For each, we report both the online reconstructed point cloud and the offline refined mesh obtained under a fixed manipulation budget of \(30\,\mathrm{s}\). Qualitative and quantitative results are shown in Fig.~\ref{fig:qual_all_objects} and Tab.~\ref{tab:quant_results_all_objects}, respectively.

Reconstruction accuracy is evaluated using the F-score metric, where \(F@\tau\) denotes the harmonic mean of precision and recall within a distance tolerance \(\tau\) (higher is better). As summarized in Tab.~\ref{tab:quant_results_all_objects}\textbf{(a)}, AURORA achieves strong reconstruction performance under a fixed \(30\,\mathrm{s}\) budget. The online point-cloud reconstruction already reaches high accuracy, with an average \(F@10\) of \(0.9671\) and \(F@5\) of \(0.8249\), showing that active reorientation can efficiently acquire informative observations. Offline mesh refinement further improves the reconstruction quality, increasing the mesh-level \(F@5\) to \(0.8624\) while preserving a high \(F@10\) of \(0.9672\). 
As shown in the Tab.~\ref{tab:quant_results_all_objects}\textbf{(b)}, AURORA also consistently outperforms non-active rotation baselines, including single-axis rotations along the \(x\)-, \(y\)-, and \(z\)-axes and a predefined fixed rotation schedule. This is further supported by the qualitative temporal comparison in Fig.~\ref{fig:uncertainty}\textbf{(a)}, where AURORA reconstructs more complete geometry with fewer missing regions and faster surface coverage. These results indicate that uncertainty-driven active reorientation improves both reconstruction efficiency and final surface completeness. More detailed qualitative comparisons with the \(x\)-, \(y\)-, and \(z\)-axis baselines are provided in Appendix~\ref{app:baseline}.

We further compare our method with recent single-view 3D reconstruction methods, TRELLIS.2~\citep{xiang2025trellis2} and SPAR3D~\citep{huang2025spar3d}, as reference baselines in Tab.~\ref{tab:quant_results_all_objects}(b). Although not designed for active in-hand reconstruction, these methods provide a useful reference for reconstruction from incomplete visual observations. For a well-defined comparison, we use the corresponding clean object image in Fig.~\ref{fig:qual_all_objects}\textbf{(a)} as input for each object. Because single-view methods do not recover metric scale, their raw outputs can have arbitrary sizes; we therefore align their mesh to the ground truth scale before computing F-scores. Thus, these scores reflect scale-normalized shape similarity, not directly usable metric reconstructions accuracy. Even after removing scale errors in their favor, our method still achieves higher mean F-scores at the \(5\,\mathrm{mm}\) and \(10\,\mathrm{mm}\) thresholds. Moreover, these baselines require clean, high-resolution inputs and may produce incomplete or non-watertight meshes under hand occlusion, limiting their suitability for metric-aware robotic applications. Detailed comparisons and qualitative results are provided in Appendix~\ref{app:single_view_comparison}.

\begin{table*}[t]
\centering
\footnotesize
\setlength{\tabcolsep}{2pt}
\renewcommand{\arraystretch}{1.05}
\caption{Quantitative results after a fixed \(30\,\mathrm{s}\) budget. 
\(F@\tau\) is the harmonic mean of precision and recall at tolerance \(\tau\). 
Single-view meshes are evaluated only with Mesh--Mesh scores after post-hoc scale alignment to the ground-truth meshes.}
\label{tab:quant_results_all_objects}

\begin{subtable}[t]{0.49\textwidth}
\centering
\caption{Results of our active strategy across objects.}
\label{tab:quant_results_objects}
\resizebox{\linewidth}{!}{%
\begin{tabular}{l ccc ccc}
\toprule
\multirow{2}{*}{Obj.}
& \multicolumn{3}{c}{PCD--PCD (Online)}
& \multicolumn{3}{c}{Mesh--Mesh (Offline)} \\
\cmidrule(lr){2-4}\cmidrule(lr){5-7}
& \(F@2\)\(\uparrow\)
& \(F@5\)\(\uparrow\)
& \(F@10\)\(\uparrow\)
& \(F@2\)\(\uparrow\)
& \(F@5\)\(\uparrow\)
& \(F@10\)\(\uparrow\) \\
\midrule
Cube & 0.2895 & 0.9337 & 0.9977 & 0.6481 & 0.9557 & 0.9957 \\
\shortstack[l]{Corner Block} & 0.2353 & 0.7354 & 0.9303 & 0.5298 & 0.8559 & 0.9488 \\
\shortstack[l]{L-shaped Block} & 0.2306 & 0.8366 & 0.9886 & 0.4674 & 0.8367 & 0.9450 \\
Pepper & 0.1770 & 0.8321 & 0.9773 & 0.5480 & 0.8890 & 0.9913 \\
Cylinder & 0.4428 & 0.8400 & 0.9425 & 0.4035 & 0.8320 & 0.9372 \\
\shortstack[l]{Cross Block} & 0.1454 & 0.7718 & 0.9664 & 0.4601 & 0.8050 & 0.9850 \\
\midrule
\textbf{Mean} & 0.2534 & 0.8249 & 0.9671 & 0.5095 & 0.8624 & 0.9672 \\
\textbf{Std.}  & 0.1054 & 0.0679 & 0.0263 & 0.0855 & 0.0535 & 0.0262 \\
\bottomrule
\end{tabular}%
}
\end{subtable}
\hfill
\begin{subtable}[t]{0.49\textwidth}
\centering
\caption{Mean comparison with baselines.}
\label{tab:quant_results_baselines}
\resizebox{\linewidth}{!}{%
\begin{tabular}{l ccc ccc}
\toprule
\multirow{2}{*}{Method}
& \multicolumn{3}{c}{Mean PCD--PCD}
& \multicolumn{3}{c}{Mean Mesh--Mesh} \\
\cmidrule(lr){2-4}\cmidrule(lr){5-7}
& \(F@2\)\(\uparrow\)
& \(F@5\)\(\uparrow\)
& \(F@10\)\(\uparrow\)
& \(F@2\)\(\uparrow\)
& \(F@5\)\(\uparrow\)
& \(F@10\)\(\uparrow\) \\
\midrule
\(x\)-axis & 0.1735 & 0.6625 & 0.8655 & 0.3595 & 0.6211 & 0.8440 \\
\(y\)-axis & 0.1463 & 0.6074 & 0.7933 & 0.3082 & 0.5643 & 0.7498 \\
\(z\)-axis & 0.2079 & 0.7786 & 0.9228 & 0.4414 & 0.7231 & 0.8872 \\
Fixed schedule & 0.1550 & 0.7015 & 0.8995 & 0.4544 & 0.6948 & 0.8904 \\
\midrule
TRELLIS.2~\citep{xiang2025trellis2} & -- & -- & -- & \textbf{0.6538} & 0.8315 & 0.9017 \\
SPAR3D~\citep{huang2025spar3d} & -- & -- & -- & 0.5571 & 0.7386 & 0.9483 \\
\midrule
\textbf{Ours (Active)}
& \textbf{0.2534} & \textbf{0.8249} & \textbf{0.9671}
& 0.5095 & \textbf{0.8624} & \textbf{0.9672} \\
\bottomrule
\end{tabular}%
}
\end{subtable}

\end{table*}

\subsection{Comparison with Active and Non-active Baselines}
\label{subsec:exp_baseline_comparison}

\subsubsection{Comparison with Non-active Baselines}

\begin{figure*}[t]
    \centering
    \includegraphics[width=0.9\linewidth]{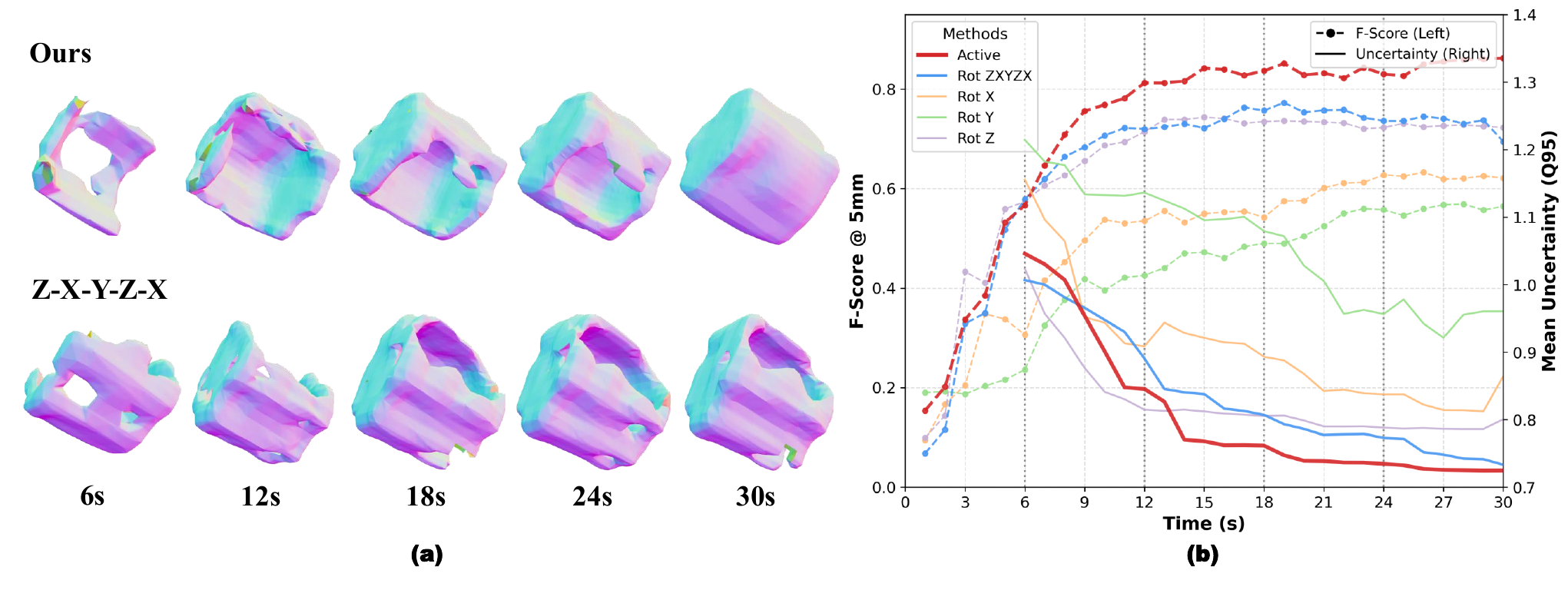}
    \caption{(a) Qualitative comparison of the reconstruction result on \textit{Cross Block} over time between our method and the Z-X-Y-Z-X baseline. (b) Online reconstruction efficiency under different rotation strategies, averaged over six objects. The curves show the \(q_{95}\) uncertainty and Mesh--Mesh F-score at the 5mm threshold within a \(30\,\mathrm{s}\) budget. 
}
    \label{fig:uncertainty}
\end{figure*}

\paragraph{Analysis of Active Reconstruction Efficiency.}
We first evaluate online reconstruction efficiency by tracking the temporal evolution of both reconstruction uncertainty and geometric accuracy under the proposed closed-loop active strategy. We use \(q_{95}\), the 95th percentile of the estimated uncertainty distribution, to indicate poorly observed, high-uncertainty regions (e.g., holes in the reconstruction). We also report the Mesh--Mesh F-score at \(5\,\mathrm{mm}\), denoted as \(F@5\), to measure the quality of the intermediate reconstructed mesh against the scanned ground truth. As shown by the red curves in Fig.~\ref{fig:uncertainty}\textbf{(b)}, our method reaches lower \(q_{95}\) and higher \(F@5\) over time, indicating effective information acquisition and improved geometric reconstruction quality. The solid red \(q_{95}\) curve decreases stepwise: when exploration begins to plateau, each replanning step (every \(6\,\mathrm{s}\), marked by vertical lines) introduces further uncertainty reduction. The increasing \(F@5\) curve further confirms that the reduced uncertainty translates into improved geometric reconstruction. After the final replanning step at \(24\,\mathrm{s}\), both curves gradually stabilize, suggesting that the most informative viewpoints have been sufficiently covered and additional observations yield diminishing returns.

\paragraph{Comparison with Open-loop Rotation Strategies.} \label{subsec:openloop}
We further compare our method against open-loop rotation schedules under the same fixed \(30\,\mathrm{s}\) manipulation budget. The dashed \(F@5\) curves in Fig.~\ref{fig:uncertainty}\textbf{(b)} show that our method improves reconstruction quality more efficiently. In particular, after the first replanning step at \(6\,\mathrm{s}\), our \(F@5\) curve grows faster than those of the baselines, indicating that the actively selected rotations provide more informative observations. Single-axis rotations perform worst, with slower F-score improvement and higher remaining uncertainty, because they fail to expose surfaces outside their limited reachable sets. The predefined multi-axis schedule performs better but remains suboptimal because it cannot adapt to the current reconstruction state. Minor non-monotonic fluctuations in the curves arise from the automatic fusion and mesh extraction pipeline, where denoising, point-cloud downsampling, and meshing can slightly change the reconstructed surface over time.

\subsubsection{Comparison with Active View-Planning Baselines}

To further evaluate active view-selection performance, we compare Ray-GPIS
with ActNeRF~\cite{actnerf} and PB-NBV~\cite{pbnbv}, adapting both methods to
our in-hand reorientation setting. We conduct 120 paired episodes in simulation
with ground-truth object poses and exact RGB--depth registration to isolate
planner performance from perception and tracking errors. We report F-AUC,
final \(F@5\), planner correlation (Corr.), and planning time. F-AUC measures
reconstruction quality over the full exploration process, while Corr. measures
the agreement between planner-estimated action scores and the actual geometric
gains of executable actions.

As shown in Table~\ref{tab:planner_ablation}(a), Ray-GPIS achieves the highest
F-AUC and final \(F@5\), indicating more efficient exploration and better final
reconstruction quality. More importantly, Ray-GPIS obtains a substantially
higher Corr. than both adapted baselines, showing that its planner scores more
reliably rank executable actions according to their actual geometric gains.
Ray-GPIS also requires substantially less planning time than ActNeRF. These
results show that the gains over open-loop strategies are not solely due to
closed-loop replanning; the proposed Ray-GPIS planner also selects more
effective reconstruction actions than alternative active view-planning methods.

\subsection{Ablation and Robustness Analysis}
\label{subsec:exp_ablation}

We further evaluate two key Ray-GPIS components through targeted ablations: (i) \emph{miss-ray interpolation}, which estimates anchors for rays without surface hits, and (ii) \emph{receptive-field (RF) integration}, which aggregates uncertainty over a local neighborhood. As shown in Table~\ref{tab:planner_ablation}(b), removing miss-ray interpolation reduces
Unseen R-AUC under finger occlusion, indicating that interpolated anchors help
the planner reason about directions whose surfaces are not directly observed.
Removing RF integration causes a substantially larger degradation in Corr.
under pose perturbations, showing that locally aggregating uncertainty makes
the action ranking considerably less sensitive to small geometric
misalignments. These results support the intended roles of the two components:
miss-ray interpolation improves reasoning over persistently unobserved regions,
whereas RF integration stabilizes uncertainty estimation under pose errors.

\begin{table*}[t]
\centering
\caption{Active view-planning comparison and targeted ablations of Ray-GPIS.
(a) Planner comparison over 120 paired simulation episodes.
(b) Component ablations under targeted failure modes.}
\label{tab:planner_ablation}

\small

\begin{minipage}[t]{0.48\textwidth}
\centering

\textbf{(a) Active view-planning comparison}\\[2pt]

\setlength{\tabcolsep}{5pt}
\begin{tabular}{lccc}
\toprule
Metric
& ActNeRF~\cite{actnerf}
& PB-NBV~\cite{pbnbv}
& \textbf{Ray-GPIS} \\
\midrule

F-AUC$\uparrow$
& $0.87{\pm}0.01$
& $0.75{\pm}0.02$
& $\mathbf{0.90{\pm}0.01}$ \\

F@5$\uparrow$
& $0.97{\pm}0.01$
& $0.85{\pm}0.02$
& $\mathbf{0.98{\pm}0.01}$ \\

Corr.$\uparrow$
& $0.16{\pm}0.03$
& $-0.36{\pm}0.07$
& $\mathbf{0.47{\pm}0.07}$ \\

Time (s)$\downarrow$
& $2.24{\pm}0.02$
& $0.39{\pm}0.02$
& $\mathbf{0.26{\pm}0.02}$ \\

\bottomrule
\end{tabular}

\end{minipage}
\hfill
\begin{minipage}[t]{0.48\textwidth}
\centering

\textbf{(b) Targeted ablations}\\[2pt]

\setlength{\tabcolsep}{6pt}
\begin{tabular}{lcc}
\toprule
Method & Score & $\Delta$ \\
\midrule

\multicolumn{3}{l}{
\emph{Finger occlusion: Unseen R-AUC$\uparrow$}
} \\

Full Ray-GPIS
& $\mathbf{0.5715}$
& -- \\

w/o miss-ray interp.
& $0.5498$
& $-0.0217$ \\

\midrule

\multicolumn{3}{l}{
\emph{Pose errors ($6^\circ$/3\,mm): Corr.$\uparrow$}
} \\

Full Ray-GPIS
& $\mathbf{0.5028}$
& -- \\

w/o RF integration
& $0.2724$
& $-0.2304$ \\

\bottomrule
\end{tabular}

\end{minipage}

\vspace{-0.1cm}
\end{table*}



\section{Limitations}

Our framework has three main limitations. First, reconstruction remains sensitive to 6D pose-tracking errors under weak visual texture or severe hand--object occlusion, motivating tactile or contact-aided tracking. Second, the fixed \(6\,\mathrm{s}\) replanning interval could be adapted based on uncertainty or tracking confidence. Finally, occasional object slip or drops highlight the need for more robust control and sim-to-real transfer.

\section{Conclusion}
\label{sec:conclusion}

{
This paper presents AURORA, an active in-hand reconstruction framework that closes the loop between online object-centric reconstruction and uncertainty-driven reorientation. At its core, Ray-GPIS estimates direction-wise reconstruction uncertainty to guide executable in-hand rotation actions under severe hand--object occlusion. Together with CAD-free 6D pose tracking and lightweight RGB-D fusion, AURORA progressively exposes under-observed regions and improves reconstruction completeness. Experiments show that AURORA outperforms non-active rotation strategies in reconstruction quality and information-acquisition efficiency, while Ray-GPIS also surpasses adapted active view-planning baselines in reconstruction performance, action-ranking quality, and planning efficiency. Targeted ablations validate the roles of miss-ray interpolation and receptive-field integration under occlusion and pose errors, and pose-perturbation tests further demonstrate robustness to moderate tracking inaccuracies. Future work will focus on improving pose-tracking robustness and in-hand manipulation stability.
}

\clearpage

\acknowledgments{
We thank the Area Chair and the anonymous reviewers for their constructive feedback and helpful suggestions.
This work was supported by the Natural Science Foundation of Shanghai
under Grant 25ZR1402370 and in part by the MoE Key Laboratory of Intelligent
Perception and Human--Machine Collaboration.
}

\bibliography{example.bib}  
\newpage
\appendix

\section*{Appendix}

\setcounter{figure}{0}
\setcounter{table}{0}
\setcounter{equation}{0}

\renewcommand{\thefigure}{A.\arabic{figure}}
\renewcommand{\thetable}{A.\arabic{table}}
\renewcommand{\theequation}{A.\arabic{equation}}

\renewcommand{\theHfigure}{appendix.A.\arabic{figure}}
\renewcommand{\theHtable}{appendix.A.\arabic{table}}
\renewcommand{\theHequation}{appendix.A.\arabic{equation}}

\section{Details of In-Hand Reorientation Policy}
\label{app:inhand_policy_details}

\subsection{Hardware Setup}
\label{app:hardware_specification}
This section summarizes the hardware used for policy training, real-world deployment, tactile sensing, and ground-truth mesh acquisition. Our hardware setup consists of an Azure Kinect DK RGB-D camera for visual sensing and a Leap Hand~\citep{shaw2023leaphand} for in-hand manipulation. Since the manipulator is equipped with tactile sensors, the tactile sensor configuration is detailed in Sec.~\ref{supp:Sensor_setting}. Policy training is conducted on a workstation equipped with four NVIDIA A40 GPUs (Sec.~\ref{supp:policy}), while real-world deployment runs on a desktop system with an Intel Core i7-13700 CPU, 32~GB of RAM, and an NVIDIA RTX 4090D GPU with 24~GB of memory. For metric computation in the main manuscript, ground-truth object meshes are captured using an EinScan Pro 2X V2 scanner.

\subsection{Policy Deployment on the Leap Hand} \label{supp:policy}

First, our system requires object to be actively manipulated. For achieving in-hand manipulation, we use the tactile-based in-hand rotation policy from \emph{Rotating without Seeing}~\citep{yin2023rotating} as the low-level reorientation controller and adapt it to the Leap Hand~\citep{shaw2023leaphand}. Since the original policy was designed for a different hand morphology (Allegro Hand) and a different tactile layout, we modify both the simulation environment and the policy interface to match our hardware.

Specifically, we replace the original hand model with the Leap Hand model and update the joint limits, actuator parameters, and fingertip geometry. To improve contact coverage during rotation, we extend the distal fingertip links, as shown in Fig.~\ref{fig:fsr_layout}. We also express commanded rotation axes in the hand-centered world frame \(W\), consistent with the main paper. The policy takes an axis command \(\mathbf{a}\in\mathcal{A}\) as input and outputs joint-level motor commands to rotate the grasped object about the commanded axis.

Before deployment, we retrain the policy in simulation to match the Leap Hand morphology and our tactile sensing setup. The reward is

\begin{equation}
R
=
w_{\mathrm{rot}}R_{\mathrm{rot}}
+
w_{\mathrm{contact}}R_{\mathrm{contact}}
+
w_{\mathrm{stable}}R_{\mathrm{stable}}
-
w_{\mathrm{act}}R_{\mathrm{act}}
-
w_{\mathrm{slip}}R_{\mathrm{slip}}
-
w_{\mathrm{drop}}R_{\mathrm{drop}}.
\label{eq:inhand_reward}
\end{equation}
Here, \(R_{\mathrm{rot}}\) encourages angular velocity about the commanded axis, \(R_{\mathrm{contact}}\) encourages sustained tactile contact, and \(R_{\mathrm{stable}}\) promotes stable object retention. The penalty terms $R_{\mathrm{act}}$, $R_{\mathrm{slip}}$, and $R_{\mathrm{drop}}$ discourage excessive actions, object slip, and dropping, respectively. This reward adaptation encourages the policy to produce reliable axis-conditioned rotations rather than merely maximizing rotation speed.

\subsection{FSR Sensor Layout and Usage} \label{supp:Sensor_setting}

\begin{figure}[htb]
    \centering
    \includegraphics[width=0.85\linewidth]{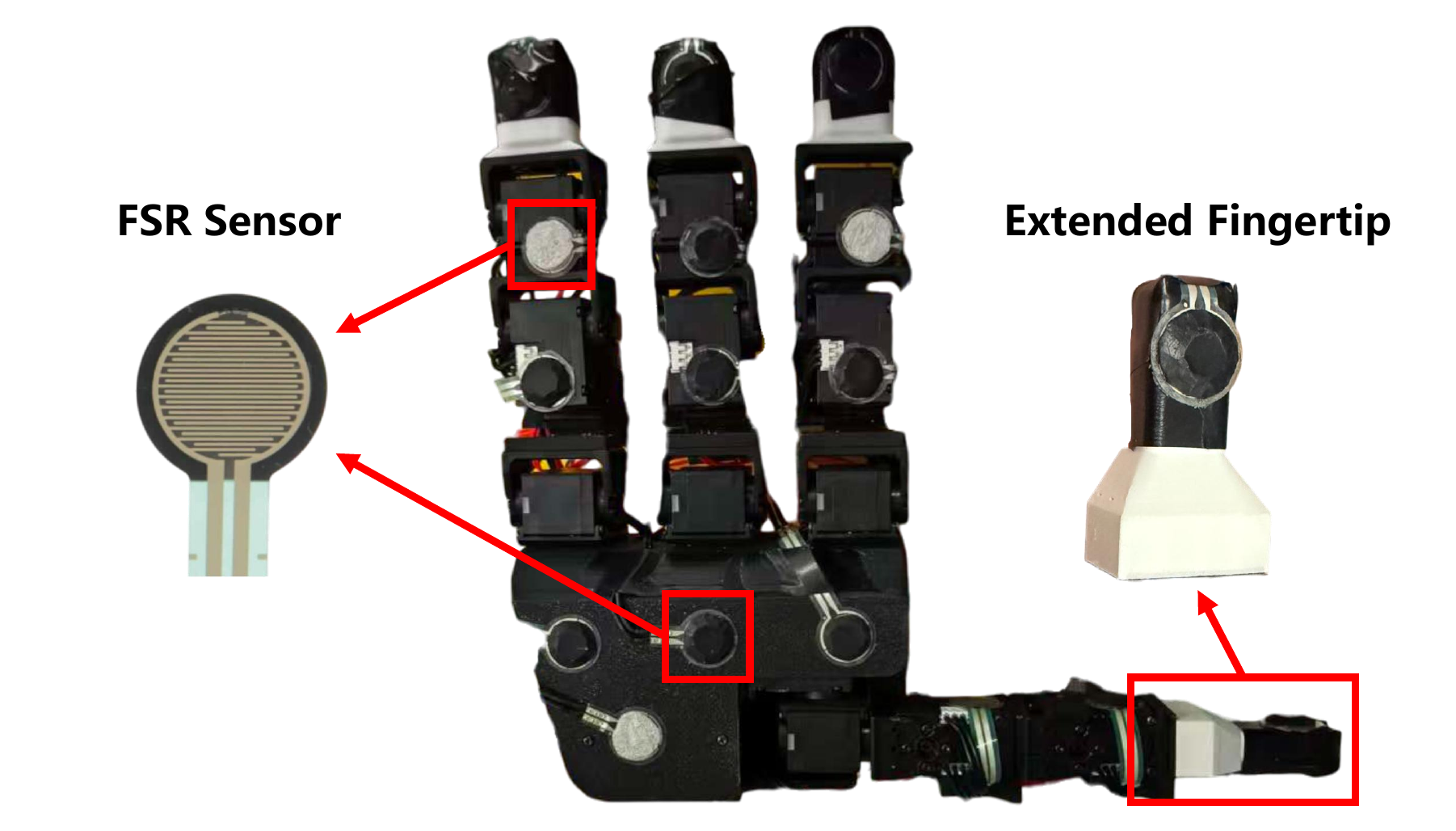}
    \caption{
    FSR sensor layout on the Leap Hand. Sensors are embedded across the fingertips, distal and middle phalanges, and the palm to provide tactile feedback for the in-hand rotation policy.
    }
    \label{fig:fsr_layout}
\end{figure}

To provide tactile feedback needed by in-hand manipulation, we mount force-sensitive resistor (FSR) sensors across the inner surfaces of Leap Hand fingertips, PIP/DIP phalanges, and palm, as illustrated in Fig.~\ref{fig:fsr_layout}. These regions correspond to the primary contact areas during in-hand rotation. FSR signals are asynchronously updated at approximately 50 Hz in a background thread,
providing the policy with the most recent contact state at each execution step.

The FSR signals are used only by the low-level in-hand reorientation policy. Although they are not used by the reconstruction pipeline, they benefit reconstruction indirectly by enabling more reliable object reorientation.

\subsection{FSR Sensor Activation and Utilization}
\label{app:fsr_utilization}

\begin{figure}[H]
    \centering
    \includegraphics[width=0.95\linewidth]{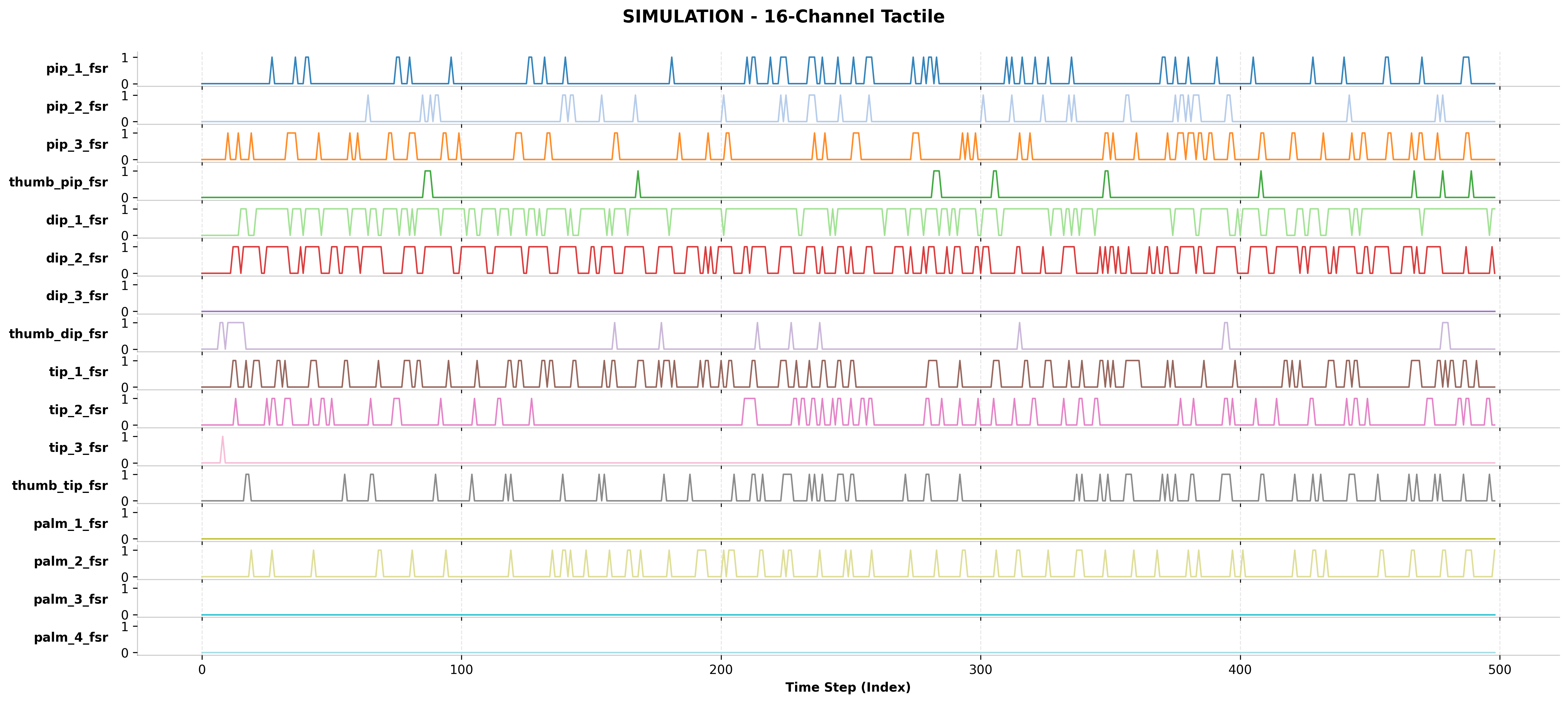}
    \vspace{0.5em}
    \includegraphics[width=0.95\linewidth]{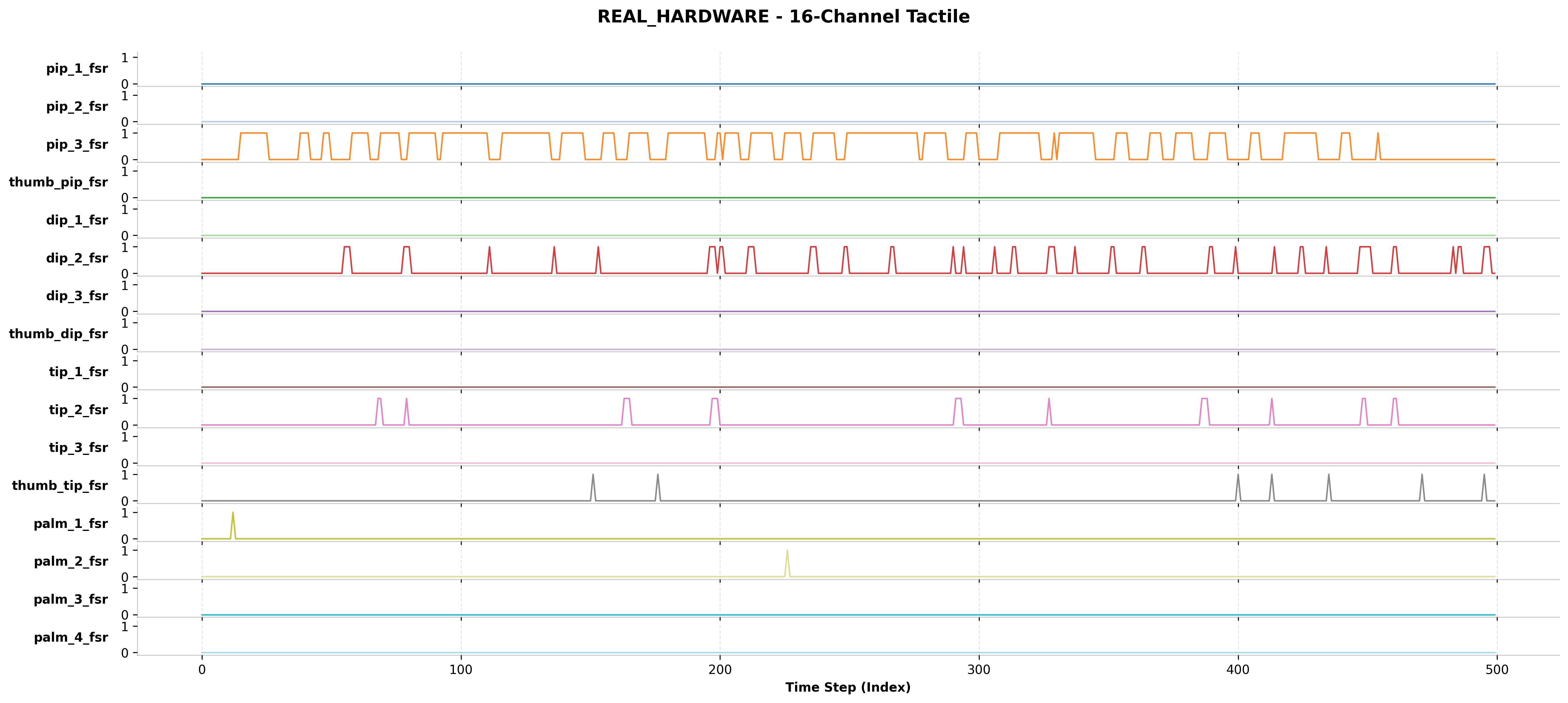}
    \caption{
    Temporal activation of the 16-channel FSR tactile signals in simulation and real-hardware deployment. 
    }
    \label{fig:fsr_utilization}
\end{figure}

To better understand how tactile feedback is used by the in-hand rotation policy, we visualize the temporal activations of the 16-channel FSR tactile signals in both simulation and real-hardware deployment. Each curve corresponds to one FSR channel, and a pulse indicates that the corresponding tactile sensor is activated by contact during manipulation.

Fig.~\ref{fig:fsr_utilization} shows that the simulated policy produces rich and distributed tactile activations across multiple fingers and palm sensors, indicating that the learned policy actively exploits contact information from different regions of the hand during object rotation. In real-hardware deployment, the activations become sparser and are concentrated on several dominant contact channels. This discrepancy is mainly caused by the physical characteristics of the real tactile sensors, including the minimum force threshold required for activation, hysteresis during activation and deactivation, and the limited stiffness of the manipulated object, which differs from the simulation setting. Despite these differences, the real system can still reliably obtain consistent tactile feedback from key contact regions, leading to successful manipulation. These observations also suggest that further improvements in sensor placement, calibration, and tactile signal normalization may  be a possible way to increase the number of informative channels and improve the stability of real-world in-hand manipulation.


\section{Details of Ray-GPIS Uncertainty Estimation}
\label{app:raygpis_details}

This section provides the mathematical details omitted from Sec.~\ref{subsec:method_active_gpis}, including hit/miss ray assignment, anchor interpolation, and GPIS posterior variance.

\subsection{Hit/Miss Ray Assignment and Anchor Interpolation}
\label{app:ray_gpis_definition}

Given the current fused point cloud \(\mathcal{P}_t\), Ray-GPIS samples a set of candidate viewing directions \(\{\mathbf{u}_i\}_{i=1}^{N}\) and casts rays from the object center \(\mathbf{c}\):
\begin{equation}
\mathbf{x}(t;\mathbf{u}_i)=\mathbf{c}+(tr)\mathbf{u}_i,\qquad t\in[0,t_{\max}],
\end{equation}
where \(r\) normalizes the ray extent. In our implementation, each ray is discretized into a set of depth samples \(\{t_{ij}\}_{j=1}^{N_t}\). We compute the minimum distance from each ray to the observed point cloud:
\begin{equation}
d_{\min}(\mathbf{u}_i)
=
\min_{j}
\min_{\mathbf{p}\in\mathcal{P}_t}
\left\|
\mathbf{x}(t_{ij};\mathbf{u}_i)-\mathbf{p}
\right\|_2 .
\label{eq:app_ray_min_dist}
\end{equation}
A ray is treated as a hit ray if this distance is below a threshold \(\tau_{\mathrm{hit}}\):
\begin{equation}
m_i
=
\mathbb{I}\!\left[
d_{\min}(\mathbf{u}_i)<\tau_{\mathrm{hit}}
\right].
\label{eq:app_hit_mask}
\end{equation}

For a hit ray, the anchor depth \(t_0(\mathbf{u}_i)\) is defined as the first depth at which the ray enters the observed surface neighborhood:
\begin{equation}
t_0(\mathbf{u}_i)
=
\min
\left\{
t_{ij}
\ \middle|\
\min_{\mathbf{p}\in\mathcal{P}_t}
\left\|
\mathbf{x}(t_{ij};\mathbf{u}_i)-\mathbf{p}
\right\|_2
<
\tau_{\mathrm{hit}}
\right\}.
\label{eq:app_first_hit_depth}
\end{equation}
The corresponding anchor point is
\begin{equation}
\mathbf{x}_0(\mathbf{u}_i)
=
\mathbf{x}(t_0(\mathbf{u}_i);\mathbf{u}_i).
\label{eq:app_anchor_point}
\end{equation}

For a miss ray, no reliable observed surface intersection is available. We therefore infer a plausible anchor depth by interpolating the depths of nearby hit-rays on the viewing sphere. Let \(\mathcal{H}_i\) denote the set of nearby hit directions:
\begin{equation}
\mathcal{H}_i
=
\left\{
j
\ \middle|\
m_j=1,\ 
\mathbf{u}_i^{\top}\mathbf{u}_j \ge \gamma
\right\},
\label{eq:app_hit_neighbors}
\end{equation}
where \(\gamma\) is an angular similarity threshold. We compute cosine-based interpolation weights:
\begin{equation}
\alpha_{ij}
=
\frac{
\exp\!\left((\mathbf{u}_i^{\top}\mathbf{u}_j-1)/\tau_{\mathrm{interp}}\right)
}{
\sum_{\ell\in\mathcal{H}_i}
\exp\!\left((\mathbf{u}_i^{\top}\mathbf{u}_{\ell}-1)/\tau_{\mathrm{interp}}\right)
},
\qquad j\in\mathcal{H}_i,
\label{eq:app_interp_weight}
\end{equation}
where \(\tau_{\mathrm{interp}}\) controls the angular interpolation bandwidth. The miss-ray anchor depth is then estimated as
\begin{equation}
t_0(\mathbf{u}_i)
=
\sum_{j\in\mathcal{H}_i}
\alpha_{ij}t_0(\mathbf{u}_j).
\label{eq:app_miss_depth_interp}
\end{equation}
If no valid neighboring hit ray exists, the ray is excluded from the current planning update. This anchor interpolation allows Ray-GPIS to evaluate uncertainty near plausible surface locations, even for directions without direct observations.

\subsection{GPIS Posterior Variance}
\label{app:train_gpis}

Ray-GPIS uses a Gaussian Process (GP) model to estimate spatial reconstruction uncertainty. Given the anchor set
\begin{equation}
\mathbf{X}
=
\left\{
\mathbf{x}_0(\mathbf{u}_i)
\right\}_{i=1}^{N_X},
\end{equation}
and the corresponding labels \(\mathbf{y}\), we model the implicit function as
\begin{equation}
f(\mathbf{x})
\sim
\mathcal{GP}
\left(
\mu_0(\mathbf{x}),
k(\mathbf{x},\mathbf{x}')
\right),
\label{eq:app_gpis_prior}
\end{equation}
where \(\mu_0(\cdot)\) is the prior mean and \(k(\cdot,\cdot)\) is the covariance kernel.

For a query point \(\mathbf{x}\), the posterior distribution is
\begin{equation}
p(f(\mathbf{x})\mid\mathbf{X},\mathbf{y})
\sim
\mathcal{N}
\left(
\mu(\mathbf{x}),
\sigma^2(\mathbf{x})
\right).
\label{eq:app_gpis_posterior}
\end{equation}
The predictive mean and variance are
\begin{equation}
\mu(\mathbf{x})
=
\mu_0(\mathbf{x})
+
\mathbf{k}_{\mathbf{x}}^{\top}
\left(
\mathbf{K}+\sigma_n^2\mathbf{I}
\right)^{-1}
\left(
\mathbf{y}-\boldsymbol{\mu}_0
\right),
\label{eq:app_gpis_mean}
\end{equation}
\begin{equation}
\sigma^2(\mathbf{x})
=
k(\mathbf{x},\mathbf{x})
-
\mathbf{k}_{\mathbf{x}}^{\top}
\left(
\mathbf{K}+\sigma_n^2\mathbf{I}
\right)^{-1}
\mathbf{k}_{\mathbf{x}},
\label{eq:app_gpis_var}
\end{equation}
where \(\mathbf{K}=k(\mathbf{X},\mathbf{X})\), \(\mathbf{k}_{\mathbf{x}}=k(\mathbf{X},\mathbf{x})\), \(\boldsymbol{\mu}_0=\mu_0(\mathbf{X})\), and \(\sigma_n^2\) is the observation noise variance.

In AURORA, we use the posterior variance \(\sigma^2(\mathbf{x})\) as a geometric reconstruction uncertainty metric. This uncertainty is high in regions less covered by the current observations and low near well-observed anchors.

\section{Details of NBV-to-Action Mapping}
\label{app:action_selection_details}

This section details how the selected NBV direction is mapped to a feasible in-hand rotation primitive. The NBV module outputs a desired viewing direction \(\mathbf{u}^{\star}_{\mathrm{NBV}}\) in the object-centered planning frame. Since the low-level reorientation policy executes rotation-axis commands in the hand-centered world frame \(W\), we first transform the NBV direction into \(W\).

Let \(T_{CO}(t)\in SE(3)\) denote the tracked object pose in the camera frame, and let \(T_{WC}\in SE(3)\) denote the calibrated camera-to-world transform. The object pose in the hand-centered world frame is
\begin{equation}
T_{WO}(t)
\triangleq
T_{WC}T_{CO}(t),
\label{eq:app_world_object_pose}
\end{equation}
with rotation component \(\mathbf{R}_{WO}(t)\). The desired viewing direction in \(W\) is then given by
\begin{equation}
\mathbf{d}_{W}
=
\frac{
\mathbf{R}_{WO}(t)\mathbf{u}^{\star}_{\mathrm{NBV}}
}{
\left\|
\mathbf{R}_{WO}(t)\mathbf{u}^{\star}_{\mathrm{NBV}}
\right\|_2
}.
\label{eq:app_desired_view_direction}
\end{equation}

We define \(\mathbf{v}_W\) as the unit vector from the current object center to the camera, expressed in \(W\). We compute the minimal rotation that aligns \(\mathbf{d}_{W}\) with \(\mathbf{v}_{W}\). The rotation angle is
\begin{equation}
\theta
=
\arccos
\left(
\mathrm{clip}
\left(
\mathbf{v}_{W}^{\top}\mathbf{d}_{W},
-1,
1
\right)
\right),
\label{eq:app_alignment_angle}
\end{equation}
where clipping improves numerical stability. When \(\mathbf{v}_{W}\) and \(\mathbf{d}_{W}\) are not parallel, the rotation axis is
\begin{equation}
\mathbf{k}
=
\frac{
\mathbf{d}_{W}\times\mathbf{v}_{W}
}{
\left\|
\mathbf{d}_{W}\times\mathbf{v}_{W}
\right\|_2
}.
\label{eq:app_alignment_axis}
\end{equation}
The corresponding minimal alignment rotation is
\begin{equation}
R_{\mathrm{err}}
=
\exp
\left(
\theta[\mathbf{k}]_{\times}
\right),
\label{eq:app_alignment_rotation}
\end{equation}
where \([\mathbf{k}]_{\times}\) is the skew-symmetric matrix of \(\mathbf{k}\).

We represent this rotation as a Lie algebra vector:
\begin{equation}
\boldsymbol{\omega}
=
\mathrm{vee}
\left(
\log(R_{\mathrm{err}})
\right)
=
\theta\mathbf{k}
\in\mathbb{R}^{3},
\label{eq:app_rotation_vector}
\end{equation}
where \(\mathrm{vee}(\cdot)\) maps a skew-symmetric matrix to its vector representation. In the main paper, we use \(\log(R_{\mathrm{err}})\) to denote this vectorized rotation representation for compactness.

Finally, we project the desired rotation vector onto the discrete action set supported by the low-level in-hand rotation policy:
\begin{equation}
\mathbf{a}^{\star}
=
\arg\max_{\mathbf{a}\in\mathcal{A}}
\mathbf{a}^{\top}\boldsymbol{\omega}.
\label{eq:app_axis_selection}
\end{equation}
This selects the feasible rotation primitive whose axis is most aligned with the desired viewpoint-correcting rotation.

For numerical robustness, when \(\left\|\mathbf{d}_{W}\times\mathbf{v}_{W}\right\|_2 < \epsilon\), we treat the two directions as nearly parallel. If \(\mathbf{v}_{W}^{\top}\mathbf{d}_{W}>0\), we set \(\boldsymbol{\omega}=\mathbf{0}\), indicating that no corrective rotation is required. If \(\mathbf{v}_{W}^{\top}\mathbf{d}_{W}<0\), we choose an arbitrary unit vector orthogonal to \(\mathbf{v}_{W}\) as \(\mathbf{k}\) and set \(\theta=\pi\). In practice, this degenerate case rarely occurs because the NBV directions are discretely sampled and the action space is coarse.

\section{Details of 6D Pose Tracking and Keyframe Selection}
\label{app:tracking_details}

This section provides additional details on the object pose tracking and keyframe selection strategy used in Sec.~\ref{subsec:method_tracking_keyframe}. AURORA uses BundleTrack~\citep{wen2021bundletrack} to estimate the object pose in the camera frame, \(T_{CO}(t)\in SE(3)\), from segmented RGB-D observations. BundleTrack maintains a pool of masked keyframes and jointly refines their poses through bundle adjustment. The resulting object poses allow depth observations from different viewpoints to be transformed into a consistent object-centered frame for fusion.

However, during in-hand manipulation, hand--object occlusions and imperfect segmentation can introduce unreliable frames. Directly inserting these frames into the keyframe pool may corrupt the fused geometry and destabilize the pose graph. We therefore apply a lightweight keyframe filter based on visibility, pose stability, and viewpoint change.

Let \(t^{-}\) denote the most recent selected keyframe. For a candidate frame at time \(t\), we define the relative motion with respect to \(t^{-}\) as
\begin{equation}
\Delta T(t)
\triangleq
T_{CO}(t^{-})^{-1}T_{CO}(t),
\label{eq:app_relative_pose}
\end{equation}
with rotation and translation components \(\Delta \mathbf{R}(t)\) and \(\Delta \mathbf{t}(t)\), respectively. The relative rotation and translation magnitudes are
\begin{equation}
\Delta\theta(t)
=
\cos^{-1}
\left(
\frac{\mathrm{tr}(\Delta \mathbf{R}(t))-1}{2}
\right),
\qquad
\Delta p(t)
=
\|\Delta \mathbf{t}(t)\|_2 .
\label{eq:app_relative_motion}
\end{equation}
We also compute the pose-update magnitude
\begin{equation}
e(t)
=
\|\log(\Delta T(t))\|_2,
\label{eq:app_pose_update_magnitude}
\end{equation}
where \(\log(\cdot)\) denotes the Lie algebra mapping from \(SE(3)\) to its tangent space.

A frame is retained as a keyframe if it satisfies both reliability and informativeness criteria:
\begin{equation}
\begin{aligned}
t \in \mathcal{K}
\Longleftrightarrow\ 
&\big(A_t\ge\tau_A\big)
\wedge
\big(e(t)\le\tau_E\big)\\
&\wedge
\max\!\left(
\frac{\Delta\theta(t)}{\tau_R},
\frac{\Delta p(t)}{\tau_T}
\right)\ge 1 .
\end{aligned}
\label{eq:app_keyframe_rule}
\end{equation}
Here, \(A_t\) denotes the visible object area measured from the segmentation mask \(\mathbf{M}_t\). The threshold \(\tau_A\) rejects frames with insufficient visible object pixels, while \(\tau_E\) filters out frames with unstable pose updates. \(\tau_R\) and \(\tau_T\) are the rotation and translation thresholds, ensuring that the selected frame provides sufficient viewpoint change relative to the most recent keyframe, either through rotation or translation. This prevents redundant frames from being inserted while preserving informative observations for multi-view fusion.

In our implementation, this filtering step is applied before the online reconstruction update. Only the selected keyframes that pass the filter \(\mathcal{K}\) are used to update the fused point cloud \(\mathcal{P}_t\). This improves the stability of both point-cloud fusion and subsequent Ray-GPIS uncertainty estimation.

\section{Details of Mesh Extraction}
\label{app:mesh_extraction_details}

AURORA represents the online reconstruction as a lightweight point cloud \(\mathcal{P}_t\), enabling low-latency uncertainty estimation and active planning. After exploration terminates, we optionally run an offline post-processing stage to obtain a watertight mesh for quantitative evaluation and visualization.

 To perform mesh reconstruction, given the final fused point cloud \(\mathcal{P}_T\), we first estimate globally consistent surface normals using FaCE~\citep{scrivener2025faraday}:
\begin{equation}
\hat{\mathbf{N}}
\triangleq
\mathcal{F}(\mathcal{P}_T),
\label{eq:app_face_normals}
\end{equation}
where \(\mathcal{F}(\cdot)\) denotes the FaCE-based normal estimation operator and \(\hat{\mathbf{N}}\) is the estimated normal field. We use FaCE because the online fused point cloud may contain non-uniform density, positional noise, and residual misalignment from pose-tracking errors. More consistent normals improve the quality of the subsequent mesh reconstruction.

We then reconstruct a watertight mesh using NKSR~\citep{huang2023nksr}:
\begin{equation}
\mathcal{M}
\triangleq
\mathcal{R}_{\mathrm{nksr}}
\left(
\mathcal{P}_T,
\hat{\mathbf{N}}
\right),
\label{eq:app_nksr_mesh}
\end{equation}
where \(\mathcal{R}_{\mathrm{nksr}}(\cdot)\) denotes the NKSR reconstruction operator. The resulting mesh \(\mathcal{M}\) is used only for final evaluation and qualitative visualization, and is not used by the online planner. During active exploration, Ray-GPIS operates directly on the online point cloud \(\mathcal{P}_t\), avoiding the latency of repeated mesh extraction.

\section{Experimental Details}
\label{app:experimental_details}

\subsection{Planner and Reconstruction Hyperparameters}
\label{app:experiment_parameters}

The main parameters used in our experiments are listed in Tab.~\ref{tab:planner_hyperparameters}.
\begin{table}[h]
\centering
\caption{Planner and reconstruction hyperparameters.}
\label{tab:planner_hyperparameters}
\begin{tabular}{l|c}
\toprule
Parameter & Value \\
\midrule
Visibility threshold \(\tau_A\) & \(0.25\) \\
Pose stability threshold \(\tau_E\) & \(0.03\) \\
Rotation threshold \(\tau_R\) & \(10^{\circ}\) \\
Translation threshold \(\tau_T\) & \(0.008\,\mathrm{m}\) \\
Ray hit threshold \(\tau_{\mathrm{hit}}\) & \(0.005\,\mathrm{m}\) \\
Angular similarity threshold \(\gamma\) & \(0.6\)\\
Angular interpolation bandwidth \(\tau_{\mathrm{interp}}\) & \(0.08\)\\
Novelty scale \(\beta\) & \(4.0\) \\
Maximum ray extent \(s_{\max}\) / \(t_{\max}\) & \(2.2\) \\
Radial band width \(\Delta s\) / \(\Delta t\) & \(0.01\) \\
Angular neighborhood \(\delta_u\) & \(10^{\circ}\) \\
Replanning interval & \(6\,\mathrm{s}\) \\
\bottomrule
\end{tabular}
\end{table}

\begin{figure}[htbp]
    \centering
    \includegraphics[width=\linewidth]{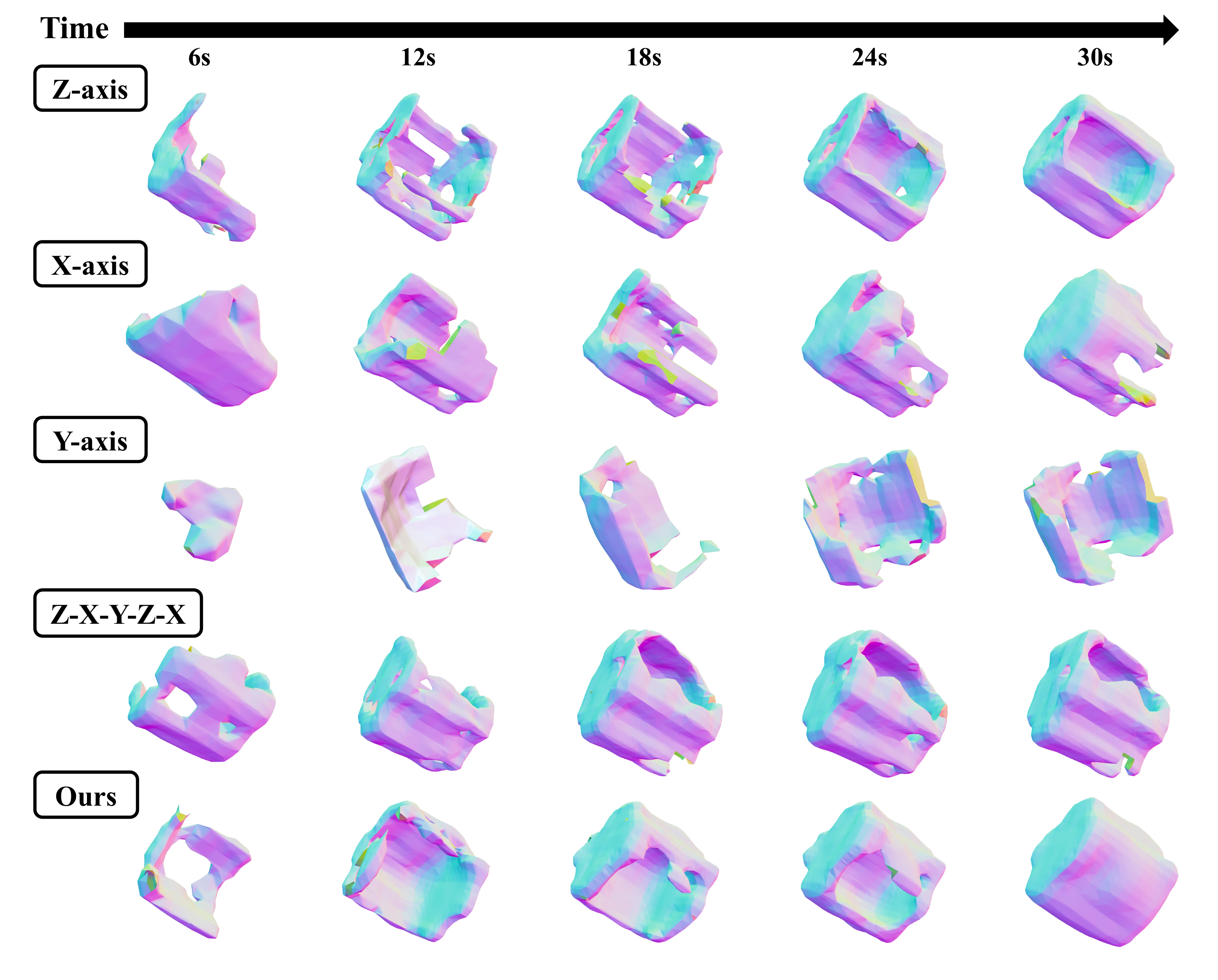}
    \caption{Qualitative comparison of reconstruction results on \textit{Cross Block} between AURORA and fixed-axis baselines.}
    \label{fig:app_baseline}
\end{figure}

\subsection{Runtime and Replanning Interval}
\label{app:runtime}

We report the runtime of the main online modules on the deployment system
described in Sec.~\ref{app:hardware_specification}. Ray-GPIS GP update,
candidate scoring, and NBV-to-action mapping require approximately
\(261\,\mathrm{ms}\), \(0.072\,\mathrm{ms}\), and \(2.42\,\mathrm{ms}\) per
planning update, respectively. Segmentation and BundleTrack together require
approximately \(256.9\,\mathrm{ms}\) per RGB-D frame. These results show that
the planner itself is sufficiently lightweight for online closed-loop
replanning.

The \(6\,\mathrm{s}\) replanning interval is determined primarily by the
low-level in-hand manipulation controller rather than planner computation.
After receiving a rotation-axis command, the hand requires approximately
\(6\,\mathrm{s}\) to complete the commanded rotation and stabilize the grasp.
Replanning earlier would therefore change the target before the previous
reorientation has been reliably executed. We consequently update the
high-level planner after each \(6\,\mathrm{s}\) manipulation interval.

\subsection{Robustness to 6D Pose-Tracking Errors}
\label{app:pose_robustness}

To evaluate sensitivity to pose-tracking errors, we manually perturb the
estimated object poses on real RGB-D sequences before reconstruction.
Specifically, we consider rotational/translational perturbations of
\(5^{\circ}/5\,\mathrm{mm}\) and \(10^{\circ}/10\,\mathrm{mm}\).
Under these perturbations, AURORA retains \(96.3\%\) and \(90.3\%\),
respectively, of its unperturbed \(F@5\) performance. These results indicate
that moderate pose errors degrade reconstruction only gradually, rather than
causing immediate failure of the reconstruction and planning pipeline.

\subsection{Additional Comparison with Open-loop Rotation Baselines}
\label{app:baseline}

Fig.~\ref{fig:app_baseline} provides additional qualitative comparisons between AURORA and the open-loop rotation baselines introduced in Sec~\ref{subsec:openloop}. Single-axis rotation strategies i.e., the \(x\)-, \(y\)-, or \(z\)-axis, provide only a limited set of viewpoints. As a result, some surface regions remain persistently occluded by the hand or are never brought into view, leading to incomplete reconstructions and visible missing areas. The predefined Z-X-Y-Z-X rotation schedule exposes more diverse viewpoints, but it remains open-loop and cannot adapt to the current reconstruction state. Therefore, some fine geometric details remain insufficiently recovered, especially in regions that require targeted viewpoint changes. In contrast, AURORA actively selects the next rotation axis based on the estimated reconstruction uncertainty, allowing it to target under-observed and high-uncertainty regions and produce more complete surfaces over time.

\begin{figure}[htbp]
    \centering
    \includegraphics[width=\linewidth]{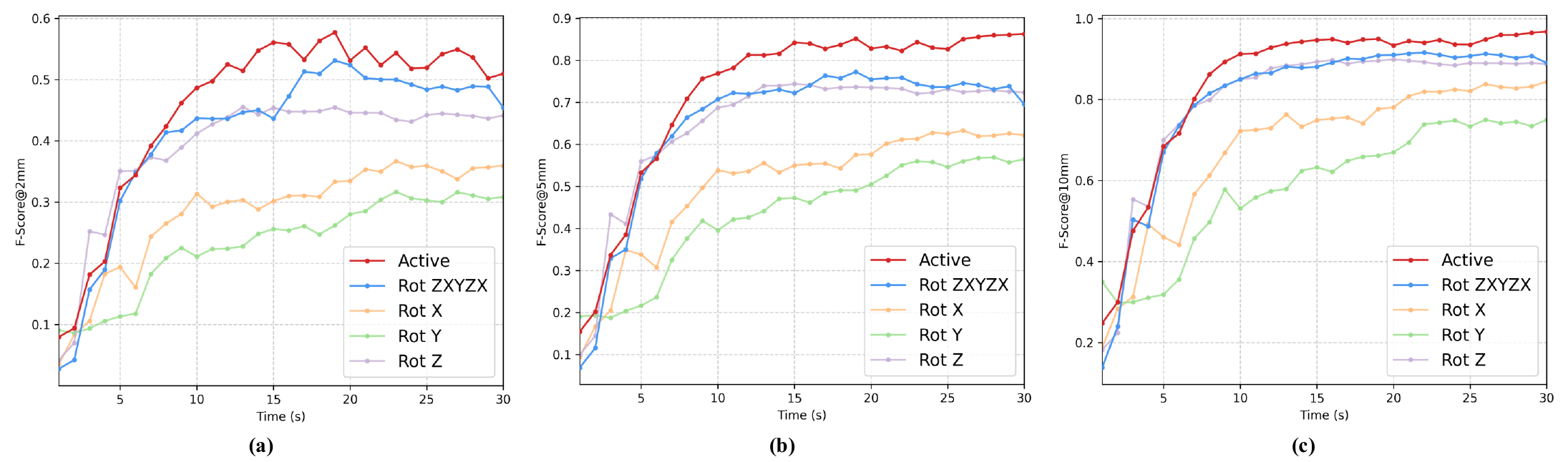}
    \caption{Temporal evolution of Mesh--Mesh F-scores for AURORA and non-active baselines under different distance tolerances: \textbf{(a)} \(2\,\mathrm{mm}\), \textbf{(b)} \(5\,\mathrm{mm}\), and \textbf{(c)} \(10\,\mathrm{mm}\), averaged over six objects.}
    \label{fig:app_temporal_fscore}
\end{figure}

We further evaluate reconstruction efficiency using the temporal Mesh--Mesh F-score curves. At each timestamp, we reconstruct a mesh from the online point cloud fused up to that time using the same post-processing pipeline and compare it with the scanned ground-truth mesh. Fig.~\ref{fig:app_temporal_fscore} reports the averaged F-scores over six objects under three distance thresholds: 2 mm, 5 mm, and 10 mm.

Because AURORA always starts with a rotation about the \(z\)-axis; therefore, its F-score curve nearly overlaps with other baselines during the first \(6\,\mathrm{s}\). After \(6\,\mathrm{s}\), the open-loop baselines improve more slowly because repeated observations along the same or predefined axes provide diminishing information gain. Although switching axes according to a predefined schedule can improve reconstruction and achieves performance close to AURORA in some trials, it is less reliable overall. For instance, object slip during in-hand manipulation can change the actual object pose, preventing the predefined schedule from consistently exposing the intended unseen surfaces. In contrast, AURORA replans from the current reconstruction state and is therefore more robust on average, maintaining faster F-score improvement and reaching higher accuracy sooner.

The curves tend to saturate after approximately \(18\,\mathrm{s}\) for two main reasons. First, for simple objects, most surfaces have already been observed by this time, leaving limited room for further F-score improvement. For more complex geometries, such as the \textit{Cross Block} shown in Fig.~\ref{fig:app_baseline}, AURORA can still fill missing mesh regions after \(18\,\mathrm{s}\). Second, in practice, the camera-to-object distance, sensor resolution, accumulated 6D tracking errors, and the downsampling and denoising operations used during point-cloud fusion all limit the mesh resolution and introduce small geometric deviations. Consequently, even our method indeed completes previously missing regions, the mesh extraction process can introduce slight distortions or misalignments elsewhere. Under fixed distance thresholds, these deviations can offset the F-score gain brought by the newly observed regions. Thus, this saturation does not indicate a failure of the active strategy.

We also note that the three single-axis baselines behave differently. This is caused not only by viewpoint coverage, but also by hardware constraints. Due to the sim-to-real gap in tactile feedback utility, contact dynamics, and friction, rotations around the \(x\)- and \(y\)-axes are less efficient and stable than the \(z\)-axis rotation and are more likely to induce object slip or unstable motions. Consequently, their reconstruction performance can be worse even under the same manipulation budget.

\subsection{Comparison with Single-View 3D Reconstruction Baselines}
\label{app:single_view_comparison}

\begin{figure}[htbp]
    \centering
    \includegraphics[width=\linewidth]{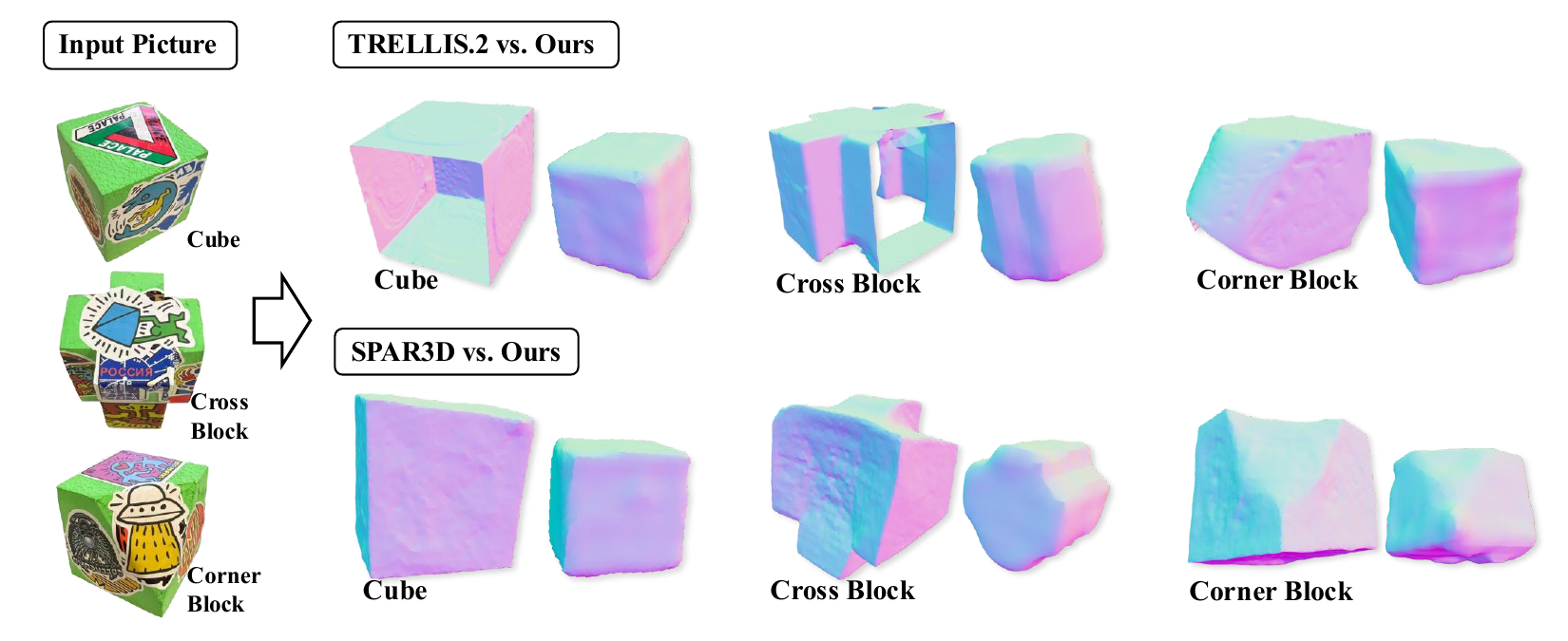}
    \caption{Visual comparison of reconstruction failure modes for single-view baselines (TRELLIS.2~\citep{xiang2025trellis2} and SPAR3D~\citep{huang2025spar3d}) versus our method on the \textit{Cube}, \textit{Cross Block}, and \textit{Corner Block}. These baseline reconstructions are generated from the high-quality input images shown on the left.}
    \label{fig:vs_trellis}
\end{figure}

To further analyze the limitations of single-view 3D reconstruction  under in-hand occlusion, we compare AURORA with two recent single-image baselines, TRELLIS.2~\citep{xiang2025trellis2} and SPAR3D~\citep{huang2025spar3d}. 
Although these methods are not designed for active in-hand reconstruction, they provide useful reference baselines because they also reconstruct 3D geometry from incomplete visual observations. 
We evaluate all methods using the Mesh--Mesh F-score at \(2\,\mathrm{mm}\), \(5\,\mathrm{mm}\), and \(10\,\mathrm{mm}\) thresholds. The quantitative results are reported in Tab.~\ref{tab:single_view_appendix}.

\begin{table}[t]
\centering
\footnotesize
\setlength{\tabcolsep}{3pt}
\renewcommand{\arraystretch}{1.12}
\caption{Mesh--Mesh reconstruction comparison with single-view 3D reconstruction baselines. \(F@\tau\) denotes the F-score under distance threshold \(\tau\), where higher is better.}
\label{tab:single_view_appendix}
\vspace{4pt}
\begin{tabular}{l|ccc|ccc|ccc}
\toprule
\multirow{2}{*}{Object}
& \multicolumn{3}{c|}{TRELLIS.2~\citep{xiang2025trellis2}}
& \multicolumn{3}{c|}{SPAR3D~\citep{huang2025spar3d}}
& \multicolumn{3}{c}{\textbf{Ours}} \\
\cline{2-10}
& \(F@2\) & \(F@5\) & \(F@10\)
& \(F@2\) & \(F@5\) & \(F@10\)
& \(F@2\) & \(F@5\) & \(F@10\) \\
\midrule
Cube
& \textbf{0.8922} & 0.9324 & 0.9579
& 0.4510 & 0.5515 & 0.8968
& 0.6481 & \textbf{0.9557} & \textbf{0.9957} \\

Corner Block
& 0.5140 & 0.6447 & 0.7833
& 0.4507 & 0.5331 & 0.8686
& \textbf{0.5298} & \textbf{0.8559} & \textbf{0.9488} \\

L-shaped Block
& \textbf{0.8124} & \textbf{0.9251} & 0.9632
& 0.5209 & 0.7244 & \textbf{0.9987}
& 0.4674 & 0.8367 & 0.9450 \\

Pepper
& 0.6612 & \textbf{0.9504} & 0.9978
& \textbf{0.6998} & 0.9459 & \textbf{1.0000}
& 0.5480 & 0.8890 & 0.9913 \\

Cylinder
& 0.5884 & 0.8861 & 0.9423
& \textbf{0.9030} & \textbf{0.9963} & \textbf{1.0000}
& 0.4035 & 0.8320 & 0.9372 \\

Cross Block
& 0.4548 & 0.6509 & 0.7660
& 0.3176 & 0.6809 & 0.9260
& \textbf{0.4601} & \textbf{0.8050} & \textbf{0.9850} \\
\midrule
Mean
& \textbf{0.6538} & 0.8316 & 0.9018
& 0.5572 & 0.7387 & 0.9484
& 0.5095 & \textbf{0.8624} & \textbf{0.9672} \\
\bottomrule
\end{tabular}
\end{table}

As shown in Tab.~\ref{tab:single_view_appendix}, our method achieves higher mean Mesh--Mesh F-scores than TRELLIS.2~\citep{xiang2025trellis2} and SPAR3D~\citep{huang2025spar3d} at the \(5\,\mathrm{mm}\) and \(10\,\mathrm{mm}\) thresholds across all evaluated objects. This suggests that actively acquiring additional observations improves metric reconstruction over single-view priors. Nevertheless, single-view methods produce relatively clean meshes, highlighting a direction for us to make future improvements.

We observe the following failure modes for the two single-view baselines in Fig.~\ref{fig:vs_trellis}. TRELLIS.2~\citep{xiang2025trellis2} preserves global structure and planar surfaces well, such as the rectangular faces of the \textit{Cube} and \textit{Cross Block}. However, it often fails to maintain topological completeness for objects with sharp edges or cutouts, producing open surfaces, incomplete cutting planes, and over-smoothed details. In contrast, SPAR3D~\citep{huang2025spar3d} tends to generate more watertight meshes but with poorer geometric alignment. It often hallucinates rear-side geometry from the front view without enforcing global consistency, causing distortions such as non-orthogonal angles and warped surfaces, as seen in the \textit{Cube} and \textit{Corner Block} examples. These failures show that single-view methods struggle to infer coherent 3D geometry from limited observations, whereas incremental viewpoint acquisition is important for both completeness and geometric fidelity.

Both baselines also share three limitations in our in-hand setting. First, their performance depends strongly on input image quality. The results in Tab.~\ref{tab:single_view_appendix} use relatively clean, high-resolution object images without dexterous-hand occlusion, whereas realistic in-hand observations often suffer from partial visibility and motion blur. Second, their meshes do not preserve metric scale. We therefore rescale each generated mesh to the corresponding ground truth before computing Mesh--Mesh metrics. This scale ambiguity prevents direct use in metric-aware downstream robotic tasks. Third, they cannot recover true object texture from occluded or unobserved regions; any missing appearance must be hallucinated from learned priors. In contrast, AURORA actively reorients the object, incrementally fuses newly observed geometry in a consistent object-centric frame, and produces metrically consistent reconstructions.

\subsection{Downstream Manipulation Evaluation}
\label{app:downstream}

Beyond geometric reconstruction metrics, we evaluate whether the reconstructed
meshes provide useful geometry for downstream manipulation. Following
\citep{gualtieri2021robotic}, we test stable object placement followed by
regrasp using meshes reconstructed by different methods. We compare two
single-view reconstruction baselines, a non-active fixed rotation schedule,
adapted active view-planning baselines, and the full Ray-GPIS pipeline.
Each method is evaluated with 30 trials per object over 10 objects.

\begin{table}[t]
\centering
\caption{Downstream manipulation success (\%), averaged over 10 objects with
30 trials per object and method.}
\label{tab:downstream_appendix}
\small
\setlength{\tabcolsep}{4pt}
\begin{tabularx}{\linewidth}{Xccc}
\toprule
Reconstruction source
& Placement$\uparrow$
& Regrasp$\uparrow$
& Placement\&Regrasp$\uparrow$ \\
\midrule

SPAR3D~\citep{huang2025spar3d}
& 30.0
& 15.0
& 5.0 \\

TRELLIS.2~\citep{xiang2025trellis2}
& 19.7
& 7.7
& 0.0 \\

Fixed schedule
& 50.0
& 44.0
& 24.0 \\

Adapted PB-NBV~\citep{pbnbv}
& 43.3
& 40.0
& 23.0 \\

Adapted ActNeRF~\citep{actnerf}
& 56.0
& 72.0
& 41.7 \\

\textbf{Full Ray-GPIS}
& \textbf{57.7}
& \textbf{77.7}
& \textbf{45.0} \\

\bottomrule
\end{tabularx}
\end{table}

As shown in Tab.~\ref{tab:downstream_appendix}, Full Ray-GPIS achieves the
highest success rates for placement, regrasp, and the complete
placement-and-regrasp task, reaching 57.7\%, 77.7\%, and 45.0\%,
respectively. It consistently outperforms the fixed schedule, the adapted
active planners, and the single-view reconstruction baselines. These results
show that the geometry recovered through uncertainty-driven active exploration
is not only more complete according to reconstruction metrics, but also more
useful for downstream placement and regrasping.

\subsection{Comparison with NeuralFeels \& Discussions}
\label{app:neuralfeels_comparison}

NeuralFeels~\citep{suresh2024neuralfeels} is closely related to our work, as both methods reconstruct object geometry during in-hand manipulation. However, their task formulations differ, making a controlled direct comparison difficult. 
Specifically, NeuralFeels uses a perceptually open-loop in-hand rotation policy that focuses on rotating the object along a fixed axis and reconstructs geometry from both visual and tactile interaction sequences to compensate for occlusion. 
In contrast, our method focuses on closed-loop planning: it actively selects the next in-hand reorientation based on current reconstruction uncertainty to observe visually occluded regions, but does not use tactile sensing.

\begin{figure}[ht]
    \centering
    \includegraphics[width=0.85\linewidth]{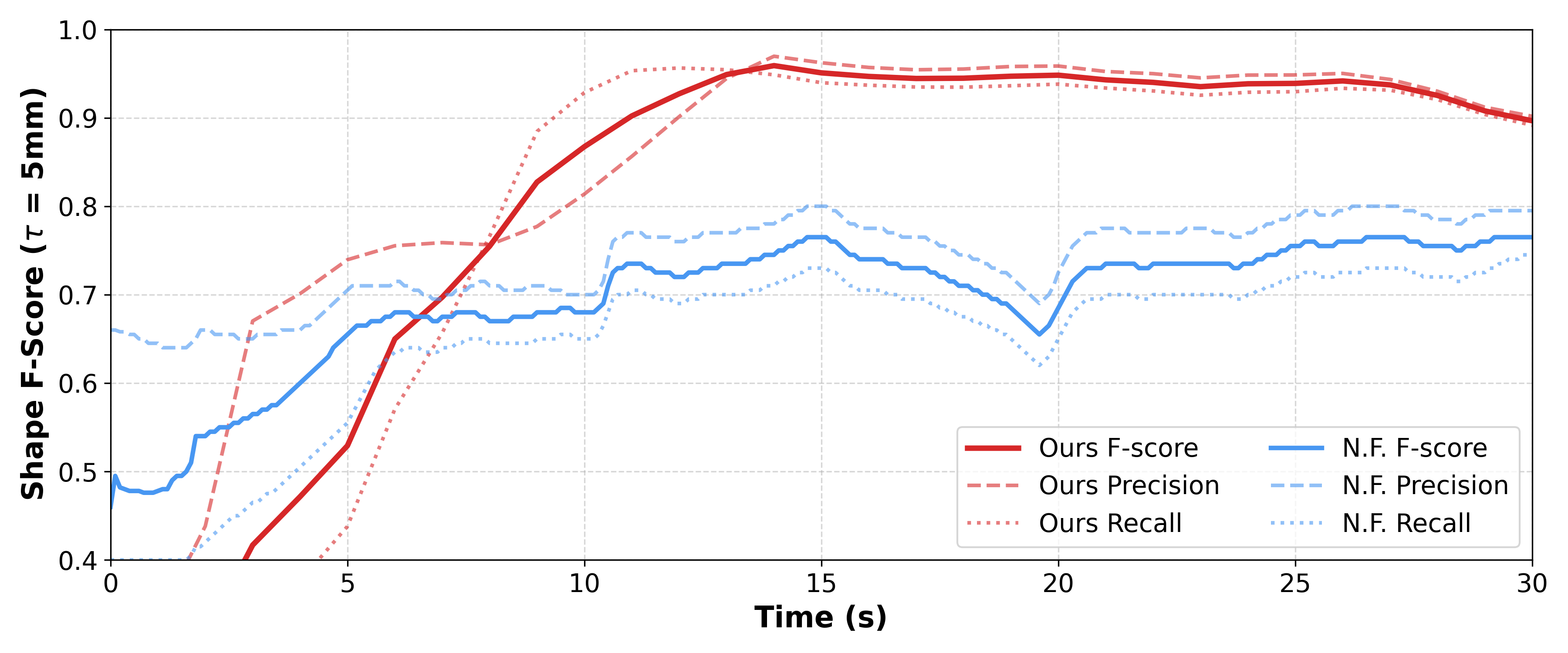}
    \caption{Quality metrics on \textit{Pepper}, comparing AURORA with NeuralFeels~\citep{suresh2024neuralfeels}.}
    \label{fig:vs_suresh}
\end{figure}

Due to differences in hardware, tactile sensing, and manipulation policies, we cannot directly reproduce NeuralFeels on our platform. We therefore conduct a limited comparison on the common \textit{Pepper} object, for which NeuralFeels provides released recordings. Both methods are evaluated under a matched \(30\,\mathrm{s}\) interaction budget using the Mesh--Mesh F-score. Results are shown in Tab.~\ref{tab:neuralfeels_comparison}, with additional metric curves in Fig.~\ref{fig:vs_suresh}. The NeuralFeels results are reproduced using the authors' released code and data without customization; we plot corresponding F-scores for comparison.

\begin{table}[t]
\centering
\footnotesize
\setlength{\tabcolsep}{6pt}
\renewcommand{\arraystretch}{1.12}
\caption{Limited comparison with NeuralFeels on the common \textit{Pepper} object under a matched \(30\,\mathrm{s}\) interaction budget. \(F@\tau\) denotes the Mesh--Mesh F-score, where higher is better.}
\label{tab:neuralfeels_comparison}
\vspace{4pt}
\begin{tabular}{lc}
\toprule
Method & \(F@5\) \\
\midrule
NeuralFeels~\citep{suresh2024neuralfeels} & 0.76 \\
Ours & 0.89 \\
\bottomrule
\end{tabular}
\end{table}

AURORA achieves an \(F@5\) score of \(0.89\), while NeuralFeels achieves an average score of \(0.76\) under the same \(30\,\mathrm{s}\) interaction budget. This comparison is subject to differences in hardware, tactile sensing, manipulation policies, task focus, and object instances; therefore, the higher score should not be interpreted as evidence that AURORA is strictly superior to NeuralFeels. In particular, we acknowledge that NeuralFeels can track textureless objects, whereas our approach requires sufficient visual texture and may fail under weakly textured conditions. The observed performance gap may also be partly attributable to object appearance: our pepper object has a similar shape but richer texture than the one shown in the NeuralFeels video. Thus, this result does not necessarily imply superior reconstruction quality of our framework.

Nevertheless, this case study suggests two potential practical benefits of our design in this setting. First, active uncertainty-guided reorientation improves reconstruction efficiency (refer to the slope rate of curves): our planner explicitly selects actions that expose under-observed regions, whereas NeuralFeels does not use reconstruction uncertainty to choose the next in-hand action. In addition, our multi-axis in-palm reorientation provides more diverse viewpoints than the fingertips' twisting motions used in NeuralFeels, which may leave the bottom surface and other occluded regions insufficiently observed. Second, our pipeline is more computationally efficient. For the same \(30\,\mathrm{s}\) observation sequence, NeuralFeels requires \(417.13\,\mathrm{s}\) of wall-clock reconstruction time, whereas our pipeline takes \(71.82\,\mathrm{s}\), including mesh extraction. This efficiency comes mainly from using a lightweight point-cloud representation for online fusion and planning instead of optimizing a dense neural SDF.

\end{document}